\documentclass{ieeeaccess}

\usepackage{cite}
\usepackage{amsmath,amssymb,amsfonts}
\usepackage{algorithmic}
\usepackage{graphicx}
\usepackage{textcomp}
\usepackage{array}
\newcommand{\supref}[1]{\expandafter\ifx\csname supref@#1\endcsname\relax S?\else\csname supref@#1\endcsname\fi}
\IfFileExists{supref.tex}{\expandafter\def\csname supref@tab:probnoattack\endcsname{S1}
\expandafter\def\csname supref@tab:probattack\endcsname{S2}
\expandafter\def\csname supref@tab:probearly\endcsname{S3}
\expandafter\def\csname supref@tab:vdelay\endcsname{S4}
\expandafter\def\csname supref@tab:wanmeas\endcsname{S5}
\expandafter\def\csname supref@tab:wanclf\endcsname{S6}
\expandafter\def\csname supref@tab:wansens\endcsname{S7}
\expandafter\def\csname supref@tab:swarm\endcsname{S8}
\expandafter\def\csname supref@tab:socket\endcsname{S9}
\expandafter\def\csname supref@tab:drrsteady\endcsname{S10}
\expandafter\def\csname supref@tab:drrburst\endcsname{S11}
\expandafter\def\csname supref@tab:kernelsteady\endcsname{S12}
}{}
\newlength{\NW}
\newcolumntype{N}{>{\centering\arraybackslash}m{\NW}}
\usepackage{url}
\def\BibTeX{{\rm B\kern-.05em{\sc i\kern-.025em b}\kern-.08em
    T\kern-.1667em\lower.7ex\hbox{E}\kern-.125emX}}

\begin{document}

\history{This work has been submitted to the IEEE for possible publication. Copyright may be transferred without notice, after which this version may no longer be accessible.}

\doi{to be assigned}

\title{Verifiable Agentic Traffic Class: A Policed Quality-of-Service Mechanism for Agentic AI Traffic, Evaluated from Simulation to Real-Kernel Adversarial Testing}
\author{\uppercase{Naveen Babu Narayanan Nair} \textnormal{and}
\uppercase{Sajay Kurup}}

\address{Comcast Corporation, Philadelphia, PA, United States}
\corresp{Corresponding author: Naveen Babu Narayanan Nair (e-mail: naveenbabunarayanannair@gmail.com). This work was not funded by an external grant or sponsor.  ORCID: Naveen Babu Narayanan Nair, 0009-0001-1425-498X; Sajay Kurup, 0009-0004-8695-7563.}

\markboth
{Nair \headeretal: Verifiable Agentic Traffic Class}
{Nair \headeretal: Verifiable Agentic Traffic Class}

\begin{abstract}
Software agents built on large language models generate traffic unlike human web traffic, yet routers have no class for it and cannot verify a sender's claim to be an agent. This paper proposes the Verifiable Agentic Traffic Class (VATC), which combines a reserved Differentiated Services Code Point (DSCP) value declaring agent traffic, a router-side classifier that demotes flows contradicting the declaration to a class served after best-effort traffic, and a rate-limited priority queue. VATC was compared with static Differentiated Services (DiffServ), a dedicated class that trusts the marking, and fq\_codel, a per-flow fair-queuing scheduler, in two simulators and on a real Linux kernel. VATC matched DiffServ under steady load. In a constructed burst it cut mean agent queuing delay by 98\% relative to DiffServ in simulation (assuming an immediate verdict), at a 12\% increase in human delay; most of the gain came from the separate class. Under a spoofed flood in the kernel, the rate limit held human latency 11 times below DiffServ's. Without verification, DiffServ and a tightly configured fq\_codel gave legitimate agents lower latency than VATC; a live four-feature classifier then demoted all 20 bulk-shaped spoofers and gave agents 0.30 ms, but missed six spoofers that copied the agent pattern. With emulated network delay, a timing feature of the classifier stopped separating agent from human flows. In real agent harnesses on a congested link, router queueing took 13\% to 63\% of task time, depending on harness and assumed model latency. The traffic was synthetic or sandbox-scale.
\end{abstract}

\begin{keywords}
Agentic artificial intelligence, Differentiated Services, fair queuing, machine learning, network scheduling, priority queuing, quality of service, spoofing detection, traffic classification.
\end{keywords}

\titlepgskip=-21pt

\maketitle

\section{Introduction}

\PARstart{S}{oftware} agents built on large language models (LLMs) plan multi-step tasks, call external tools, and exchange messages with models and with other agents. In this paper, \emph{agentic traffic} means the network traffic these agents produce, in particular the chains of dependent request and response exchanges between an agent, the inference services it calls, and the tools it invokes. An \emph{artificial intelligence (AI) gateway} is a middleware component, such as a proxy, service, or library, through which an organization channels and mediates such calls \cite{infoq}. Cisco's 2026 wide-area network (WAN) analysis reports that AI inference flows differ measurably from ordinary web flows in duration, rate, and direction, and projects that AI inference could account for about a quarter of network traffic by 2035 \cite{cisco}. These figures come from one vendor's measurements and model and describe inference flows rather than agent sessions as such; Section~III-A gives them with their limits.

Two problems follow. First, standard router quality-of-service (QoS) mechanisms have no class for agentic traffic, so a latency-sensitive agent call shares queues with web and bulk transfers (Section~II). Second, any new priority marking invites misuse: a flow can claim agent status to obtain priority, so a marking is useful only if it can be checked.

This paper proposes the Verifiable Agentic Traffic Class (VATC) to address both problems for enterprise, campus, and customer-premises networks. VATC has three components: an enumerated Agentic QoS Identifier (AQI) mapped to a reserved Differentiated Services Code Point (DSCP) at ingress; a router-side verification engine that compares the declared marking with observed flow behavior and demotes flows that do not match; and a dedicated agentic queue that is served first but policed to a fixed rate, so that errors in verification cannot let marked traffic take the whole link.

Under steady load, VATC and ordinary Differentiated Services (DiffServ) produced the same latency. In a simulated correlated burst, VATC reduced mean agentic queuing delay by 98\% relative to DiffServ (mean of 15 seeds; an upper bound, because it assumes an immediate verdict) at the cost of a 12\% increase in the delay of the concurrent human surge. A controlled ablation attributed most of this gain to giving agent traffic a class of its own, not to the agent-specific components. The rate cap bounded the harm that a spoofed flood caused to human traffic. In the kernel, a live four-feature classifier demoted every bulk-shaped spoofed flow but not spoofers that copied the agent message pattern, and under a flood without verification DiffServ and a tightly configured \texttt{fq\_codel} gave legitimate agents lower latency than VATC. Router queuing accounted for 13\% to 63\% of task time of two real agent harnesses on a congested 20~Mbit/s link, depending on the harness and the assumed model latency (Sections~VI-D, VI-J, and VI-L).

The evaluation has three limitations. The synthetic traffic generator is parameterized from the same published statistics that motivate the design, so the results show how the mechanism behaves on traffic with that signature and not that real agent traffic has it. The classifier was scored on complete flows, and the kernel verifier was trained and tested on flows from the same generator, so its clean separation of bulk-shaped spoofers shows that the mechanism works, not that real spoofers are this easy to detect. The burst scenarios are constructed stress tests, and the model latency in the harness measurement was assumed. The conditions under which VATC offers no advantage are stated in Sections~VI-A and VII-D.

The contributions of this paper are as follows.

\begin{enumerate}
\item A design for a verifiable agentic traffic class (Section~IV) with an explicit deployment scope (edge and enterprise routers behind a trusted gateway), a specified, but not exercised, out-of-band channel for the application-layer hop-count feature, and a sub-budget probation policy for new flows.
\item A controlled attribution of the latency gain, using a factorial ablation and gain accounting that separate class isolation, scheduling rule, rate cap, and verification, on identical arrival sequences (Section~VI-D).
\item An evaluation at three levels: a simulator with 15 seeds, a separately implemented second simulator, and real Linux-kernel queuing with real Transmission Control Protocol (TCP) traffic under honest and spoofed load, including \texttt{fq\_codel} at three parameter settings, a cap-ceiling sweep, and a live inline verifier over five seeds (Sections~VI-A to VI-K).
\item A measurement, with two real agent harnesses and a mock model endpoint, of the share of agent task time spent in a router queue and how much each queue discipline recovers (Section~VI-L).
\item A quantification of the cost of probation and of the priority exposure it limits, with a disclosure of classifier accuracy on partial flows (Sections~VI-C and VI-K).
\end{enumerate}

The paper is organized as follows. Section~II reviews related work, III characterizes agent traffic, and IV presents the design. Sections~V and VI give the methodology and results, VII discusses deployment and limitations, and VIII concludes.

\section{Background and Related Work}

\textbf{Differentiated Services and Expedited Forwarding. }RFC~2474 and RFC~2475 define the DiffServ architecture and the 6-bit DSCP that routers use to select a per-hop behavior (PHB) \cite{rfc2474,rfc2475}. None of the standard codepoints is reserved for AI or agentic traffic, and link-layer priority marking such as IEEE~802.1p is likewise traffic-type agnostic. The Expedited Forwarding (EF) PHB of RFC~3246 gives low delay and jitter to premium traffic only up to a configured rate, so that the premium class cannot monopolize the link \cite{rfc3246}; the AGENTIC class proposed here follows the same rate-policed pattern. Further IETF documents cover DiffServ service-class guidelines \cite{rfc4594}, interconnection classes \cite{rfc8100}, a lower-effort PHB \cite{rfc8622}, and three-color policing markers \cite{rfc2697,rfc2698}; none defines a class for agentic traffic.

\textbf{Active queue management and fair queuing. }Controlled Delay (CoDel) is an active queue management (AQM) algorithm that keeps standing queues short by dropping or marking packets when sojourn time (the time a packet has spent in the queue) stays above a target \cite{rfc8289}. The Flow Queue CoDel (FQ-CoDel) scheduler of RFC~8290 combines CoDel with per-flow queues and gives priority to sparse flows \cite{rfc8290}. It needs no marking and no class configuration, which makes it the natural alternative to a dedicated class and the baseline against which VATC must be compared (Sections~VI-D and VI-J). Low Latency, Low Loss, and Scalable Throughput (L4S) uses modified explicit congestion notification and cooperating end hosts \cite{rfc9330}. The Linux cake queue discipline maps DiffServ code points into priority tiers with per-flow fair queuing \cite{cake}; it was not evaluated here.

\textbf{5G QoS Identifier. }The 3rd Generation Partnership Project (3GPP) defines the 5G QoS Identifier (5QI) as an enumerated set of standardized classes, each bound to a packet delay budget, an error rate, and a priority level \cite{3gpp}, and an Internet Engineering Task Force (IETF) Internet-Draft describes how to map 5QI values to DSCP at the boundary between a 5G core and a DiffServ network \cite{cbs}. The Agentic QoS Identifier of this paper uses the same structure.

\textbf{Industry and standards activity. }A telemetry vendor argues that monitoring tuned to human-scale traffic will miss agent behavior \cite{plixer}, and an industry article observes that agent-initiated outbound calls can bypass conventional gateway controls \cite{infoq}. Internet-Drafts address agent identity and authorization \cite{aip}, application protocols for agents \cite{rosenberg,agtp}, and the use of agents to manage networks \cite{hong,chuyi,jadoon}; to the authors' knowledge, none defines packet-level identification or QoS differentiation for agentic traffic, and the agent identity work \cite{aip} could serve as a trust anchor for the declared marking.

\textbf{Marking manipulation, compute-layer, and semantic approaches. }Rahman et al.\ trained a neural-network ensemble to detect DSCP manipulation from deviations in latency, jitter, and throughput relative to the declared class and reported 99.28\% accuracy on a generic dataset \cite{rahman}. This paper adopts the principle of verifying a declared marking against observed behavior, applies it to an agent-specific signature, and uses lighter classifiers suited to per-flow inference at a router. Du et al.\ propose QoS-aware token scheduling for inference quotas across agents, which allocates compute and involves no routers or packets \cite{du}; Zhang et al.\ propose semantic, intent-driven resource allocation for agent-to-agent communication \cite{zhang}, a longer-term direction; VATC is designed to work with existing DiffServ hardware. A separate body of work uses agents to operate networks \cite{xiao,pellejero,reifert}, the inverse of the problem considered here, in which agents are a workload on the network.

\textbf{Positioning. }To the authors' knowledge, no prior work defines a packet-level marking and router-scheduling mechanism for agentic AI traffic and pairs it with behavioral verification against a spoofing adversary. VATC addresses that gap.

\section{Traffic Characterization and Problem Statement}

\subsection{Reported traffic signature}

Table~\ref{tab:signature} lists the statistics reported by Cisco \cite{cisco} on which the traffic model of Section~V-B and the verification features of Section~IV-C are based. The duration, rate, and upstream statistics describe AI inference flows compared with regular web transactions; the per-task and inference-share figures come from a single example task. The report does not state that agent sessions as a whole have these flow-level properties, and the smooth packet arrival assumed by the generator is an assumption motivated by token-by-token response generation, not a reported measurement. Section~VI-F tests only whether the assumed direction of these differences appears in sessions measured on sockets.

\Table[t!](width=\textwidth){STATISTICS REPORTED BY CISCO FOR AI INFERENCE FLOWS AND REGULAR WEB TRANSACTIONS \cite{cisco}\label{tab:signature}}{\begin{tabular}{|l|r|r|}
\hline
\textbf{Characteristic} & \textbf{Regular web (non-LLM)} & \textbf{AI inference} \\
\hline
Median flow duration & 643~ms & 1,292~ms (about 2$\times$) \\
Median flow rate & 3,803~kbit/s & 338~kbit/s (about 0.1$\times$) \\
Flows with more upstream than downstream bytes & about 0.5\% & about 9\% \\
Traffic per task, agent against person & baseline & up to $+$450\% (one example task) \\
Share of that increase from inference & not applicable & about 70\% (same example) \\
\hline
\end{tabular}}

\subsection{Why queuing delay can matter for agent tasks}

A reasonable objection is that, in an agent system, model inference and tool execution dominate task latency, so router queuing is immaterial. A recent profiling study of open agent harnesses found that tool execution contributed 13\% to 48\% of end-to-end latency and harness orchestration up to a further 9\% \cite{aries}. That study did not report network time, and no published breakdown known to the authors reports the share of agent-task latency spent in router queues.

Agent tasks are chains of dependent calls, so queuing delay accumulates. If call $i$ of a task needs a service time $s_i$ (model inference, tool execution, and propagation) and waits an extra time $q_i$ in a congested router queue, the task takes $\sum_i (s_i + q_i)$ and the router's share of the task time is $\sum_i q_i / \sum_i (s_i + q_i)$. Queuing matters when $q$ is comparable to $s$: it is negligible against a call that takes tens of seconds and large against a tool call that takes tens of milliseconds. How much queuing delay agent calls encounter is an empirical matter, so it was measured rather than assumed. Section~VI-L reports the task times of two real agent harnesses whose traffic crossed a router with a link congested by bulk transfers. On that link the router queue accounted for 13\% to 63\% of task time, depending on the harness and on an assumed model latency of 0.5 to 5~s per call, and for 7\% to 13\% with a shallower buffer. The measurement supported the premise that queuing can matter for agent tasks on congested links. It does not show that VATC is needed in that setting, because static DiffServ removed almost all of the extra delay (Section~VI-L). VATC should therefore be read as relevant for congested access links and latency-sensitive agents, not as a general remedy for agent latency.

\subsection{Problem statement}

A router forwards a mixed population of flows $F = \{f_1, \ldots, f_n\}$, each belonging to an unobserved true class $c(f) \in \{\text{Human}, \text{Bulk}, \text{Agentic}\}$. A subset of flows carries a declared marking $d(f) \in \{\text{none}, \text{Agentic}\}$, which may be truthful ($c(f) = \text{Agentic}$) or adversarial ($c(f) \neq \text{Agentic}$, declared to obtain priority). The router must select, per packet and at line rate, a forwarding treatment $t(f)$ that satisfies three requirements. First, genuinely agentic flows that declare themselves receive bounded, low queuing delay, consistent with the targets an operator would bind to an AQI (Section~IV-B). Second, declared but unverified flows cannot consume unbounded link capacity, whether through legitimate bursts or adversarial marking. Third, the collateral cost to other traffic classes is bounded and small, even under adversarial or correlated-burst conditions. No existing DSCP class, 5QI-style enumeration, or IETF draft defines $t(f)$ for this population.

\section{Proposed Mechanism: Verifiable Agentic Traffic Class}

VATC has three components: an enumerated marking scheme (the AQI), a router-side verification engine, and a policed-priority scheduler. Fig.~\ref{fig:arch} shows the arrangement. An agent framework or gateway at the network edge supplies an AQI value, which is mapped to a DSCP. The ingress router classifies on the DSCP. Traffic without an AQI marking goes directly to the existing best-effort (BE) and scavenger (SCAV, a low-priority class for bulk transfers) classes, while AQI-marked traffic is verified. Verified flows are served through a rate-capped priority queue, and flows that fail verification are demoted to a PENALTY class below best-effort. New flows without a verdict yet wait in a small probation (PROB) lane (Section~IV-D).

\Figure[t!](topskip=0pt, botskip=0pt, midskip=0pt)[width=0.92\textwidth]{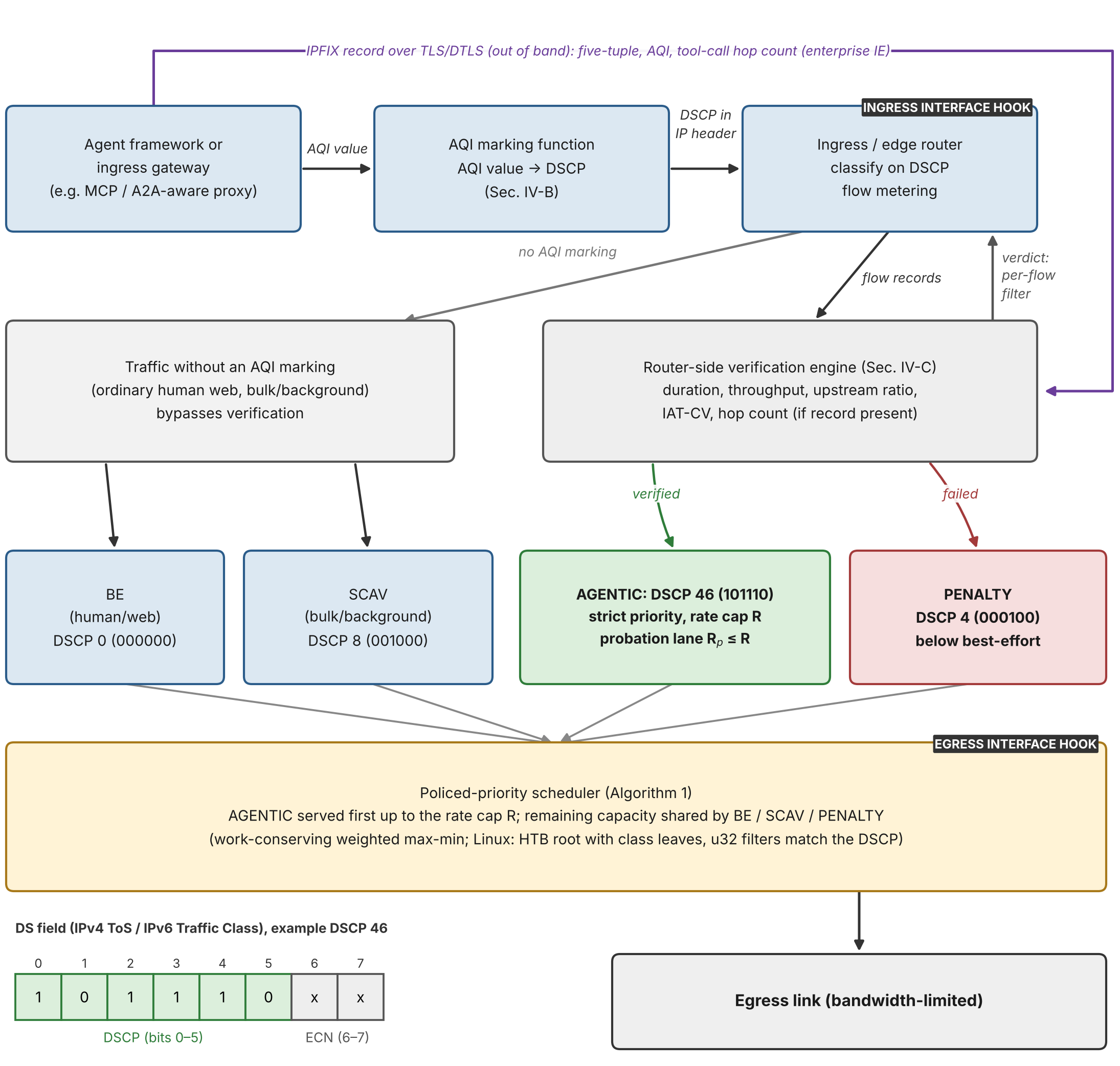}
{VATC architecture with protocol markings. DSCP values and bit patterns are the testbed stand-ins (46 coincides with the standard Expedited Forwarding code point; a deployment would use a reserved pool). The inset shows the differentiated services (DS) field (IPv4 Type of Service, IPv6 Traffic Class): DSCP in bits 0--5, explicit congestion notification (ECN) in bits 6--7. Flow metering sits at the ingress interface hook; DSCP classification and scheduling sit at the egress interface hook. The hop-count record travels out of band from the gateway to the verification engine.\label{fig:arch}}

\subsection{Deployment scope and trust model}

VATC is specified for enterprise, campus, and customer-premises equipment (CPE) access routers, and for gateway-fronted aggregation points. These devices forward in software or in an extensible data plane such as extended Berkeley Packet Filter (eBPF) programs, and the operator also controls the ingress gateway. VATC is not specified for core routers built on application-specific integrated circuits (ASICs) and forwarding at hundreds of gigabits per second. This boundary is stated because the verification engine and the application-layer feature described below depend on it.

Responsibilities are divided between two planes. The data plane performs only operations that DiffServ hardware already performs at line rate: matching the DSCP, selecting a class queue, and enforcing the AGENTIC rate cap. The verification engine runs outside the forwarding path on per-flow records and installs its verdicts as per-flow actions, namely a filter on the flow five-tuple that moves a failed flow to the PENALTY class. No packet payload is inspected, and the router never decrypts Transport Layer Security (TLS) payloads.

\textbf{Verdict latency. }Where the verifier runs sets how quickly a verdict arrives. Three placements are possible. In the first, the verifier runs on the router itself and reads the router's own flow table, as a software or eBPF data plane allows, so no export step exists. In the second, the router exports flow records with IP Flow Information Export (IPFIX) to a collector next to it, which runs the verifier and installs verdicts through a control channel. In the third, a central collector serves many routers. The experiments of Sections~VI-C and VI-K model the first placement, with the verdict available after the $N$-th packet, and Section~VI-K adds a delay between the $N$-th packet and the verdict to represent the other two.

The reaction time is the sum of the observation window (the first $N$ packets of a flow, the probation window of Section~IV-D), the export interval, the inference time, and the verdict-installation time. IPFIX specifies no timer values: an exporter expires a flow after a period of inactivity and may export long-running flows at regular intervals \cite{rfc5470}. The flow-cache defaults of Cisco Flexible NetFlow are an inactive timeout of 15~s and an active timeout of 1800~s \cite{ciscofnf}, so an exporter at its defaults may report a short flow only after 15~s of idleness, and a long flow after up to 30 minutes; operators can shorten the timers at the cost of more export load. Inference is fast by comparison: scoring one flow took 0.7~$\mu$s with logistic-regression weights applied in plain Python and 82~$\mu$s through scikit-learn, and a 200-tree random forest took 17~$\mu$s per flow in a batch of 1,000 (8.4~ms one flow at a time), measured on a cloud processor and not on router hardware. Installing a verdict took about 0.5~ms per filter on a Linux traffic-control data plane (median 0.49 to 0.52~ms for \texttt{u32} and \texttt{flower} five-tuple filters, 99th percentile at most 1.2~ms apart from one 6.6~ms outlier), the same as the 0.49~ms needed to start the \texttt{tc} process itself, and it did not grow with 0, 1,000, or 4,000 filters already installed; the measurement used an idle virtual interface in a virtual machine and no hardware offload. The export interval therefore dominates whenever the verifier sits behind an exporter.

With the verifier on the router, the verdict arrives soon after the $N$-th packet. Behind an exporter with default timers, a flow that lasts seconds ends before its verdict is installed and verification cannot demote it. In the delay measurement of Section~VII-C, the median agentic connection lasted 0.5 to 3.1~s, depending on the round-trip time (RTT), and the 90th percentile 3.3~s, so the exporter-based placements can act on agent flows only if their timers are set to about a second. Section~VI-K measures how the benefit of the lane shrinks as the delay grows. For this reason the rate cap and the probation sub-budget, and not the classifier, are the primary safeguards in VATC: they bound the capacity that any marked traffic can take while the verdict is pending, and verification reduces the residual exposure for longer flows.

\subsection{Agentic QoS Identifier}

The AQI is an enumerated class family analogous in structure to the 5QI \cite{3gpp}. Each AQI value binds a packet delay budget, a jitter budget, and a priority level to a category of agentic workload, for example interactive tool-call round trips or long-running background agent tasks. AQI values are mapped to a reserved pool of DSCP codepoints at the network edge by an agent framework, by a gateway aware of agent protocols such as the Model Context Protocol (MCP) or the Agent2Agent (A2A) protocol, or by an ingress classifier, in line with existing practice for mapping 5QI to DSCP \cite{cbs}. The testbed used DSCP~46 for AGENTIC, 8 for SCAV, 4 for PENALTY, and 0 for BE as stand-in values; a deployment would allocate codepoints from a pool reserved for experimental or local use \cite{rfc2474}. The evaluation used a single AQI class, which is sufficient to exercise the mechanism. A full enumeration, such as one class for latency-sensitive interactive agents and one for background agent traffic, is future work.

\subsection{Router-side verification engine}

Every AQI-marked flow is subject to continuous behavioral verification, not a one-time admission check, although the experiments applied a single verdict per flow, after its first $N$ packets. The verifier uses five per-flow features: flow duration, throughput, the ratio of upstream to downstream bytes, the coefficient of variation of packet inter-arrival time (IAT-CV), which captures how smooth the arrival process is, and the tool-call hop count. The first four follow the signature in Table~\ref{tab:signature}. The hop count is an addition: it is the number of hops in the agent's tool-call chain, which a genuine multi-step agent exhibits and a naive spoofer without a multi-agent integration does not.

\textbf{Transport of the hop-count feature. }The hop count is an application-layer quantity, consumed once per flow and not per packet, so it need not travel in the data plane. In the specified design, an ingress gateway that terminates or observes the agent session at the application layer, for example an MCP- or A2A-aware proxy, exports a metadata record for each AQI-marked flow to the verification engine. The record contains the flow five-tuple, the AQI value, the observed tool-call chain depth, and the session start time. It is carried in IPFIX using an enterprise-specific Information Element, which the IPFIX information model supports \cite{rfc7011,rfc7012}, over a mutually authenticated TLS or Datagram TLS (DTLS) channel, as RFC~7011 provides. The data plane sees only the DSCP. Table~\ref{tab:record} gives the record layout. The gateway exports the record when it first observes the flow and exports it again whenever the chain depth changes, so the verifier always holds the latest value.

\Table[t!](width=\textwidth){PROPOSED GATEWAY-TO-VERIFIER IPFIX RECORD. THE NUMBER IN PARENTHESES IS THE IANA INFORMATION ELEMENT IDENTIFIER. THE TWO ENTERPRISE-SPECIFIC ELEMENTS ARE PROPOSED NAMES AND ARE NOT REGISTERED\label{tab:record}}{\begin{tabular}{|l|l|r|l|}
\hline
\textbf{Field} & \textbf{Information element} & \textbf{Bytes} & \textbf{Meaning} \\
\hline
Source address & sourceIPv4Address (8) or sourceIPv6Address (27) & 4 or 16 & Flow source \\
Destination address & destinationIPv4Address (12) or destinationIPv6Address (28) & 4 or 16 & Flow destination \\
Source port & sourceTransportPort (7) & 2 & Flow source port \\
Destination port & destinationTransportPort (11) & 2 & Flow destination port \\
Protocol & protocolIdentifier (4) & 1 & Transport protocol \\
Session start & flowStartMilliseconds (152) & 8 & Time the gateway first saw the flow \\
AQI value & agenticQosIdentifier (enterprise) & 1 & AQI declared for the flow \\
Chain depth & toolCallChainDepth (enterprise) & 1 & Tool-call hops observed so far \\
\hline
\end{tabular}}

The data fields of one record occupy 23 bytes for IPv4 and 47 bytes for IPv6, before the IPFIX message and set headers of RFC~7011. The export rate equals the rate at which new AQI-marked flows appear at the gateway; as an arithmetic example, not a measurement, 1,000 new flows per second with IPv6 records would need about 0.4~Mbit/s of record data on the channel. The record rate, the export latency, and the time to install a verdict as a filter were not measured.

In-band carriers were considered and are not the primary design. IPv6 Hop-by-Hop options are unreliable at line rate because routers may skip them or process only a subset \cite{rfc9673}, TCP options have limited space and are visible to end hosts, and Geneve or VXLAN metadata is appropriate only where the operator already runs an overlay. If the gateway record is missing, verification falls back to the four network-observable features. The record channel was not exercised in the experiments: in the simulator the hop count is a per-flow attribute available to the classifier, which corresponds to a gateway that has already delivered its record.

\subsection{Probation for new flows}

A newly marked flow cannot be verified before the verifier has observed it, so a policy is needed for its first packets. Full priority invites micro-burst theft, in which an adversary opens many short flows that each end before verification; holding them in best-effort penalizes legitimate short agent sessions; and buffering them until a verdict exists would delay every new flow and need per-flow buffer memory. 

VATC therefore uses a sub-budget probation lane. For its first $N$ packets, a flow declared AGENTIC is served in a strict-priority lane that holds at most $R_p$ packets and is served at up to $R_p$ packets per tick. The lane is drawn from the AGENTIC cap $R$ (with $R_p < R$), so aggregate probationary traffic cannot exceed $R_p$ however many short flows are opened. The count $N$ covers marked packets that carry a payload in the declared direction, so the handshake and pure acknowledgments do not use up the probation, and Section~VI-C shows that three packets are the smallest window that gives a stable decision. A probationary packet that arrives when the lane already holds $R_p$ packets is re-marked into best-effort, as a policer with remarking would do; this can cause mild reordering within a flow, which was not measured. 

From packet $N{+}1$ on, the flow is served in its verdict class, AGENTIC if verified and PENALTY otherwise, provided the verdict has been installed. If it has not, as can happen when verdicts depend on IPFIX export intervals of seconds, the flow stays in the probation lane until the verdict arrives, so that aggregate exposure remains bounded by $R_p$ however slow the verdict, at the cost that legitimate flows share the lane for longer. Section~VI-K uses a verdict available at packet $N{+}1$, the software data-plane case, as its baseline, then adds a verdict delay to represent a verifier behind a flow exporter, and evaluates $N \in \{3, 5, 10, 20\}$. Continuous re-verification applies after the verdict. A limit on the rate at which one marking source can create new probationary flows would reduce the exposure further; it is specified but was not evaluated.

\textbf{Flow state. }Verification and probation need per-flow state at the router or at the co-located verifier: the five-tuple key, a packet counter for the probation window, the time of the first packet, running sums from which the four timing and volume features are computed, the latest gateway record, and the verdict (pending, verified, or penalized). Because the features come from running sums, the state has a constant size per flow. State is created on the first marked packet and removed after an operator-set idle timeout; a flow that resumes after removal is treated as new and re-enters probation, which the bound $R_p$ already covers. When the table is full, a new marked flow receives no probation priority and is served as best-effort, so exhausting the table cannot enlarge the priority lane. A verdict that demotes a flow is installed as a five-tuple filter and removed with the flow state. The state size, timeout, and table capacity are deployment parameters, and none was measured.

\subsection{Policed-priority scheduler}

Verified agentic traffic is served first, up to a hard per-tick rate cap $R$, in the manner of the EF PHB \cite{rfc3246}. Flows that fail verification are demoted to a PENALTY class below best-effort. The remaining capacity is shared among BE, SCAV, and PENALTY by weighted max-min fairness, a work-conserving rule: capacity that a short queue cannot use is redistributed to the classes that can. Capping the class bounds the cost that any combination of legitimate burst and residual verification error can impose on the rest of the link. Algorithm~1 summarizes one scheduling tick.

\begin{figure}[t!]
\noindent\fbox{\begin{minipage}{0.96\columnwidth}\small
\textbf{Algorithm 1: Per-tick VATC scheduling}\par\smallskip
\ttfamily\raggedright
\textbf{Input:} queues Q per class, capacity C, agentic cap R, probation depth R\_p, weights w\_k\\
1. a $\leftarrow$ min(R, $|$Q[AGENTIC]$|$, C); serve a from Q[AGENTIC]; C $\leftarrow$ C $-$ a\\
2. p $\leftarrow$ min(R\_p, $|$Q[PROB]$|$, R $-$ a, C); serve p from Q[PROB]; C $\leftarrow$ C $-$ p\\
3. A $\leftarrow$ \{BE, SCAV, PENALTY\} with backlog\\
4. while C $>$ 0 and A is not empty:\\
\hspace*{1.2em}5. share(k) $\leftarrow$ C $\cdot$ w\_k / $\sum_{j \in A}$ w\_j\\
\hspace*{1.2em}6. serve each k with backlog $\le$ share(k) in full; remove k from A; reduce C\\
\hspace*{1.2em}7. if no class was removed: serve floor(share(k)) from each k, give the rounding remainder to the highest-weight classes; C $\leftarrow$ 0
\end{minipage}}
\end{figure}

\section{Evaluation Methodology}

\subsection{Simulator and policies}

The primary tool is a discrete-time queuing simulator, driven by synthetic flow records and written in Python (NumPy, pandas, and scikit-learn), that represents one shared router link serving several traffic classes. It was used instead of ns-3 or a hardware testbed because the simulator lets the scheduler, classifier, and arrival process be varied together; ns-3 or hardware validation is future work, and real-kernel experiments complement the simulator (Section~V-H). Every policy within a scenario was run on a single shared arrival realization for each seed, so differences between policies could not be caused by different random traffic. Three policies were used in the main comparison:

\begin{itemize}
\item Best-effort: one first-in, first-out (FIFO) queue with no differentiation.
\item Static DiffServ: existing DSCP classes only. No codepoint is reserved for agentic traffic today, so agentic flows fall into the default class with ordinary web traffic, and bulk traffic uses a scavenger class with weight 0.3 relative to best-effort's 1.0. This models current practice. Its weights were not tuned for the burst; a more aggressive static configuration, such as a dedicated priority class for agent traffic, is represented by the trusted-marking policies of Section~V-F.
\item VATC: as in Section~IV, with a rate-policed AGENTIC class ($R = 300$ packets per tick) served first, a PENALTY class (weight 0.15) for flows that fail verification, SCAV for bulk (weight 0.3), and BE (weight 1.0) for everything else.
\end{itemize}

All policies use the work-conserving scheduler of Algorithm~1. The ablation of Section~V-F adds policies that vary one factor at a time, and Section~V-G adds fair-queuing baselines.

\subsection{Synthetic traffic population}

Each run generates 6,000 synthetic flows in four classes, chosen to approximate a near-term adoption mix: human and web (55\%), bulk and background (25\%), genuinely agentic (15\%), and adversarial flows that declare an agentic marking without being agentic (5\%). The statistics in Table~\ref{tab:signature} parameterize the feature distributions of the agentic class. The adversarial class has two types: naive spoofers, which mislabel ordinary human or bulk traffic (45\% of adversarial flows), and adaptive spoofers (55\%), which imitate the agentic duration, throughput, and arrival smoothness to evade detection. Neither type can reproduce the hop-count feature except by chance, modeled as an 8\% probability of an accidental multi-hop value. The reference population (seed 42) contained 3,345 human, 1,482 bulk, 886 genuinely agentic, and 287 adversarial flows. Each flow was expanded into packet arrivals with a gamma-distributed inter-arrival process whose shape was set to match the flow's target IAT-CV, giving smooth (low-CV) or bursty (high-CV) arrivals as appropriate.

\subsection{Verification classifier}

Logistic regression and a depth-6, 200-tree random forest were trained on the five features of all flows that declare an agentic marking, with a binary label (genuinely agentic or adversarial), using a stratified 70/30 split. The model with the higher area under the receiver operating characteristic curve (ROC-AUC, shortened to AUC below) drove the router simulation. Features were computed over complete flows (Section~VI-C reports accuracy on partial flows).

\subsection{Congestion regime}

An under-utilized link that never queues makes all policies equal, and a permanently overloaded link makes differentiation trivial. The link capacity was set to 1,500 packets per tick. The reference population offered about 1.26 million packets against a capacity of 750,000 over the 500 arrival ticks, an oversubscription of about 1.7, which represents a sustained congestion event on an access or aggregation link. The simulation ran for 500 ticks of arrivals plus a 200-tick drain period. Packets not served in this window (about 206,000 per seed under every policy, because the link is oversubscribed) were excluded from latency statistics and counted against the completion rate. Ticks are an abstract unit with no assigned clock rate: values are comparable between policies but not convertible to the milliseconds of the kernel tests. The AGENTIC rate cap $R$ is 300 packets per tick (20\% of link capacity) in the simulator and 60 of 200~Mbit/s (30\%) in the kernel tests.

\subsection{Correlated-burst scenario}

Beyond steady load, a stress scenario was constructed in which 400 genuinely agentic flows (an agent swarm reacting to one trigger) and 1,500 human flows (a concurrent human surge caused by the same trigger) arrive at one tick, with each flow's duration compressed to 1--3 ticks to concentrate the load. The scenario was not calibrated against measured data: no production telemetry on the frequency, size, or timing of such events was available, and the scenario should be read as a constructed stress test. It stands in for a higher-adoption point on Cisco's projected trajectory \cite{cisco} without being calibrated to it. Where agentic traffic remains a small and sparse share of load, the steady-state result (Section~VI-A) is the more representative one.

\subsection{Adversarial flood and ablation design}

The attribution study (Section~VI-D) adds to the burst a flood of 50 spoofed flows, declared agentic and assumed to have evaded verification, each sending at an equal share of a total rate of $m \times R$ packets per tick for 20 ticks from the burst tick, with $m \in \{0, 0.5, 1, 2, 4\}$. Policies varied three factors: whether declared flows are verified or the marking is trusted; whether the AGENTIC class uses strict priority or a generalized-processor-sharing (GPS) weight of 3 relative to best-effort (GPS divides capacity among classes in proportion to their weights); and whether the class is capped at $R$ or uncapped. Every policy sees the same arrival realization at a given seed and flood level, with the same verification outcomes and FIFO order within each class.

The allocation rules of the four scheduling variants use link capacity $C$ per tick, AGENTIC queue length $q_A$, cap $R$, and class weights $w_{\mathrm{AGENTIC}} = 3$, $w_{\mathrm{BE}} = 1$, $w_{\mathrm{SCAV}} = 0.3$, and $w_{\mathrm{PEN}} = 0.15$. \emph{Capped strict priority} (VATC) serves $\min(R, q_A, C)$ packets of the AGENTIC class first, and the remaining capacity goes to the other classes by weighted max-min fairness. \emph{Uncapped strict priority} serves $\min(q_A, C)$ first. \emph{Uncapped GPS} places AGENTIC in the weighted max-min computation of Algorithm~1 with weight 3 and no cap, so that when all four classes are backlogged it is entitled to $C \cdot 3/4.45 \approx 1{,}011$ of $C = 1{,}500$ packets and otherwise receives $\min(q_A, \text{share})$. \emph{Capped GPS} limits the AGENTIC demand to $\min(q_A, R)$. In every variant a class never receives more than it has queued, unused capacity is redistributed, and FIFO order is kept within each class, so the 1,011-packet entitlement is a bound that applies only while all classes are backlogged, not the amount that AGENTIC is served.

\subsection{Fair-queuing baselines}

Two baselines address the question whether per-flow fairness removes the need for a dedicated class. In the simulator, per-flow fair queuing gave every flow its own queue and an equal share of capacity (the fair-queuing half of FQ-CoDel, without packet drops). In the real kernel, an \texttt{fq\_codel} policy was attached below a Hierarchical Token Bucket (HTB, a Linux shaper that enforces a rate and a ceiling for each class) root shaped to the link rate, with the leaf limit set to 1000 packets. Three parameter settings were run: the default (target 5~ms, interval 100~ms), a tuned setting (1~ms and 20~ms), and a tight setting (200~$\mu$s and 5~ms).

\subsection{Real-kernel testbeds}

Two harnesses apply the actual Linux traffic-control (tc) mechanisms, HTB with strict-priority classes and DSCP matching through u32 filters, to real TCP traffic from a client and server written with the Python standard library. A Mininet topology had a router node applying the shaped configuration on its server-facing egress interface, with three client subnets (agentic, human, bulk). A lighter network-namespace (netns) harness applies the same configuration across two namespaces joined by a virtual Ethernet pair. 

In both, static DiffServ was two HTB classes under a root shaped to the link rate: best-effort (rate 1/1.3 of the link) and scavenger (0.3/1.3, matched on DSCP 8), each with a Stochastic Fairness Queuing (SFQ, round robin over hashed per-flow buckets) leaf and allowed to borrow up to the link rate, so that agentic, human, and spoofed traffic all share the best-effort class. VATC adds an AGENTIC class matched on DSCP 46 with strict priority (HTB priority 0), rate equal to ceiling at the cap, and a 1,000-packet first-in, first-out leaf, and a PENALTY class matched on DSCP 4, with best-effort, scavenger, and penalty weighted 1.0, 0.3, and 0.15 over the remaining capacity. Best-effort was a single 1,000-packet FIFO. The links added no propagation delay (RTT below 1~ms) and the sockets used CUBIC congestion control; a third setup, with client, router, and server in three namespaces and \texttt{netem} delay on the router-to-server link, was used only for the measurement of Section~VII-C. Both ran on a 200~Mbit/s link with the AGENTIC class capped at 60~Mbit/s, about 50\% above the 40~Mbit/s peak concurrent demand of 180 burst sessions. Steady phases used 60 sessions per archetype and burst phases 180 sessions opened within about 150~ms. 

The spoofed swarm (Section~VI-I) was 20 bulk-shaped sessions declared AGENTIC. Its size is the number of spoofed sessions that would pass a classifier with the held-out miss rate (0.345) if 180 legitimate sessions were accompanied by spoofed sessions at the adversarial-to-agentic ratio of the synthetic population (287/886), $\mathrm{round}(180 \times 287/886 \times 0.345)$; verification was off in these tests. 

A live verifier replaced this proxy in one experiment (Section~VI-I): a process in the client namespace that reads every packet on the shaped interface through a raw packet socket and tracks each TCP flow whose first packet carries the AGENTIC marking. After eight data packets it computes four features (log duration, log throughput, upstream byte share, and IAT-CV), scores them with a logistic-regression model trained offline, and moves a flow scored as non-agentic to the PENALTY class with a tc filter matching its source port. The hop count is not used, because the harness has no gateway. Until the verdict a flow stays in the AGENTIC class (optimistic probation); the sub-budget lane of Section~IV-D is not implemented in the kernel. Packet times are kernel timestamps taken at the tap, after the egress queue. The model was trained on three runs (seeds 101 to 103) and tested on five (seeds 42 to 46), with legitimate agentic flows as the positive class and bulk-shaped spoofed flows as the negative class; six adaptive spoofers that copied the agent message pattern were present in the training runs but were excluded from training.

\subsection{Metrics}

Per class and policy, the metrics were mean and 95th-percentile (p95) packet queuing latency and completion rate (the fraction of offered packets served in the window). For the classifier, they were accuracy, precision, recall, F1, and ROC-AUC on the held-out split.

\subsection{Multi-seed protocol}

Sections~VI-A to VI-D report a reference run (seed 42) so that all tables and consistency checks refer to the same population, except where stated. The pipeline (flow generation, classifier training, steady-state simulation, burst simulation, and ablation) was then repeated for 15 independent seeds (42 to 56). Means were reported with 95\% confidence intervals computed from the $t$ distribution with 14 degrees of freedom (multiplier 2.145) over the 15 seeds. Seed 42 was the development seed and is one of the 15. The kernel experiments of Sections~VI-I and VI-J are separate harness runs, so the same configuration can differ between them (legitimate agentic latency under plain VATC with the flood: 20.6, 20.1, and 19.9~ms).

\subsection{Real-harness task-time measurement}

To measure how much of an agent task is spent in a router queue, two agent harnesses were run through a Linux network-namespace testbed with an agent host, an edge router, and a server. Both router interfaces were shaped to 20~Mbit/s with HTB, the base RTT was 40~ms (netem), and the router queue was a 1,000-packet FIFO (about 0.6~s at that rate), or 250 packets in one sensitivity run. Harness A was mini-swe-agent \cite{miniswe,sweagent} (version 2.4.6, unmodified), whose model calls crossed the network and whose bash tool ran locally. Harness B was a tool-calling loop written with the OpenAI Python client and the MCP Python SDK \cite{mcpsdk} (version 2.3.0), in which model calls and remote tool calls both crossed the network. The model endpoint was a mock server returning a scripted trajectory of 4 to 10 tool steps per task (about 8 model calls), fixed by the task identifier and a seed, so that every router condition ran the same 24 tasks. The model latency was an assumed parameter, with a median of 0.5, 2, or 5~s per call (lognormal, $\sigma = 0.35$); the experiment measures the delay that calls incur in the router, not model inference. Traffic used plain Hypertext Transfer Protocol (HTTP) without streaming.

Six router conditions were compared. In the uncongested condition there was no background load. In the congested condition two bulk uploads, two bulk downloads, and four web-like users shared the single FIFO queue. The DiffServ condition used the same load with the static DiffServ configuration of Section~V-H (bulk demoted to the scavenger class, agent traffic sharing the best-effort class with the web). The VATC condition used the same load with the VATC data plane of Section~V-H (policed strict-priority AGENTIC class at 5~Mbit/s with a 200-packet leaf, DSCP 46 set at the edge with nftables, verification bypassed because every agent flow was legitimate, and no PENALTY class, which is not exercised when all flows are honest; the other classes as in Section~V-H). Two further conditions used the same load with Linux \texttt{fq\_codel} as the only discipline, with no DSCP classes (an HTB root at the link rate with an \texttt{fq\_codel} leaf of 1,000 packets): the default setting and the tuned setting (target 1~ms, interval 20~ms) of Section~V-G. They were run after the other four conditions, with the same tasks and load, at model latencies of 0.5 and 2~s. Task time was split into model processing, tool processing, and network time (call time minus server-side processing). The router-queue share of task time was the paired difference between the congested and uncongested task times divided by the congested task time, computed over the same task identifiers; 95\% intervals are percentile bootstrap intervals over tasks, pooled over two repeats (48 paired tasks per row). Tasks in a cell share one link, so the intervals are optimistic. Harness B used a 300~s server keep-alive, because idle disconnects otherwise aborted tasks during long model calls.

\section{Results}

\subsection{Steady-state performance}

Table~\ref{tab:steady} gives steady-state latency and completion rate by class and policy for the reference run. Under best-effort, all classes degraded similarly (mean latency 179 to 185 ticks, completion 78.5\% to 81.0\%), because one shared queue protects no class. Under static DiffServ and VATC, bulk traffic was isolated in the scavenger class and was the main casualty (mean latency 266.0 ticks, completion 71.8\% under both), while human and genuinely agentic traffic saw near-zero latency and completed in full.

\Table[t!](width=\textwidth){STEADY-STATE LATENCY (TICKS) AND COMPLETION RATE, REFERENCE RUN (SEED 42)\label{tab:steady}}{\setlength{\NW}{2.0cm}\setlength{\tabcolsep}{4pt}\begin{tabular}{|l|l|N|N|N|}
\hline
\textbf{Policy} & \textbf{Class} & \textbf{Mean} & \textbf{p95} & \textbf{Completion} \\
\hline
Best-effort & Human & 179.207 & 356 & 81.0\% \\
Best-effort & Bulk & 184.961 & 355 & 78.5\% \\
Best-effort & Agentic (legitimate) & 181.033 & 352 & 78.6\% \\
Static DiffServ & Human & 0.007 & 0 & 100.0\% \\
Static DiffServ & Bulk & 266.042 & 389 & 71.8\% \\
Static DiffServ & Agentic (legitimate) & 0.006 & 0 & 100.0\% \\
VATC & Human & 0.000 & 0 & 100.0\% \\
VATC & Bulk & 266.039 & 389 & 71.8\% \\
VATC & Agentic (legitimate) & 0.007 & 0 & 100.0\% \\
\hline
\end{tabular}}

\textbf{Under steady load, VATC offered no advantage over static DiffServ in this simulation.} The reason is structural. Genuinely agentic flows are 15\% of the flow count but only about 1\% of packets in the reference population, because their per-flow rate is low. Moving bulk traffic to a scavenger class, which DiffServ already supports, leaves enough spare capacity for agentic traffic to avoid queuing whether or not it has its own class. The steady-state case therefore does not motivate VATC; the burst case does.

\subsection{Correlated-burst scenario}

Table~\ref{tab:burst} reports the reference run of the correlated agent-swarm and human-surge scenario (Section~V-E), and Fig.~\ref{fig:burst} shows the 15-seed means. Under best-effort the swarm and the surge are indistinguishable and collapse together. Static DiffServ improves on this because the scavenger class removes bulk contention, but agentic packets still share a queue with the human surge (agentic mean 23.6 ticks). VATC reduced agentic mean latency to 0.47 ticks, a 98.0\% reduction relative to DiffServ and 99.8\% relative to best-effort, and reduced agentic peak latency from 38 to 5 ticks and drain delay from 43 to 9 ticks. The cost to the concurrent human surge was bounded: relative to DiffServ, mean latency rose by 12.1\%, peak latency by 7.9\%, and drain delay by 7.0\%. Bulk traffic was equally affected under both differentiated policies, which is the intended consequence of isolating bulk in the scavenger class.

\Table[t!](width=\textwidth){CORRELATED BURST, REFERENCE RUN (SEED 42): AGENTIC SWARM, HUMAN SURGE, AND BULK TRAFFIC (TICKS). AG. = AGENTIC; HUM. = HUMAN; DRAIN = TIME FOR THE CLASS TO COMPLETE AFTER THE BURST\label{tab:burst}}{\setlength{\NW}{1.8cm}\setlength{\tabcolsep}{4pt}\begin{tabular}{|l|N|N|N|N|N|N|N|}
\hline
\textbf{Policy} & \textbf{Ag. mean} & \textbf{Ag. peak} & \textbf{Ag. drain} & \textbf{Hum. mean} & \textbf{Hum. peak} & \textbf{Hum. drain} & \textbf{Bulk mean} \\
\hline
Best-effort & 248.8 & 264 & 269 & 244.6 & 266 & 273 & 167.9 \\
Static DiffServ & 23.6 & 38 & 43 & 19.0 & 38 & 43 & 221.4 \\
VATC & 0.5 & 5 & 9 & 21.3 & 41 & 46 & 221.5 \\
\hline
\end{tabular}}

\Figure[t!](topskip=0pt, botskip=0pt, midskip=0pt)[width=4.8in]{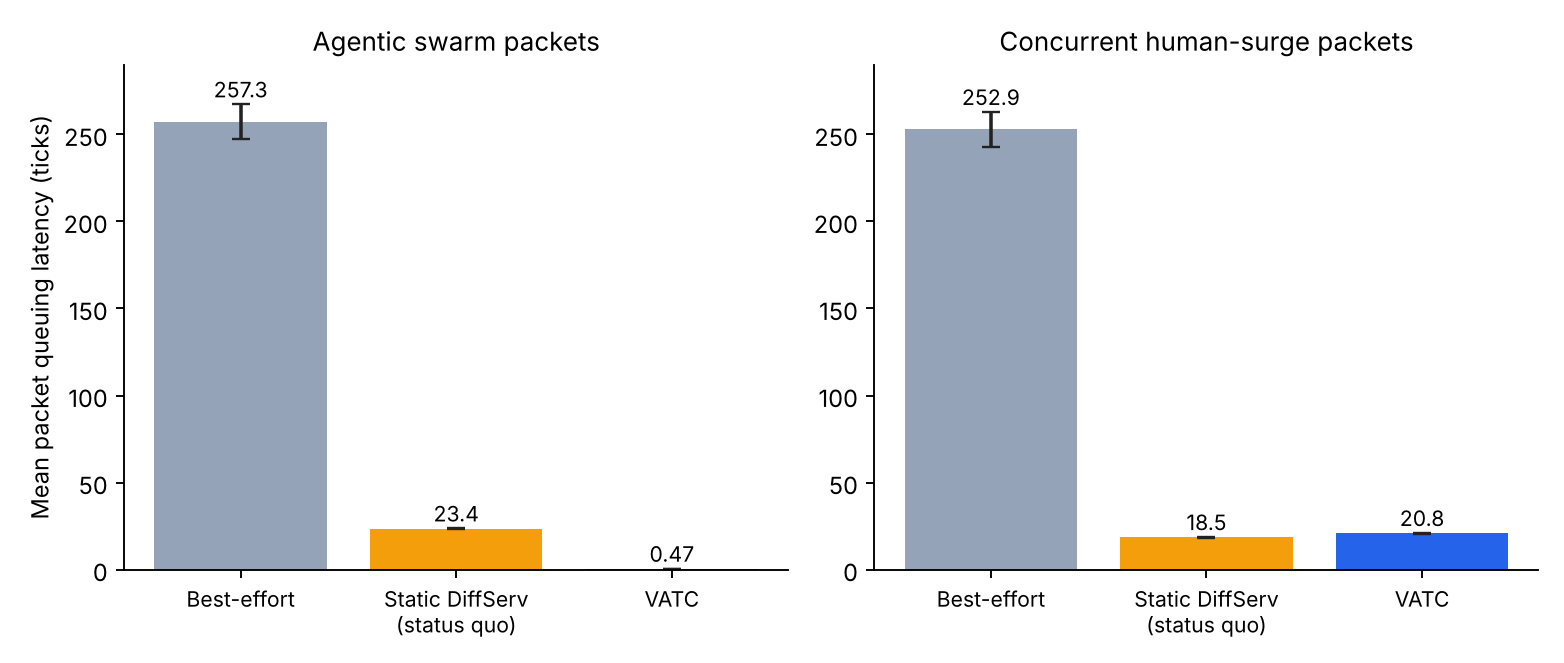}
{Mean queuing latency in the correlated burst, mean of 15 seeds with 95\% confidence intervals.\label{fig:burst}}

\subsection{Verification classifier}

Table~\ref{tab:classifier} reports classifier performance on the held-out split of the reference run, and Fig.~\ref{fig:confusion} shows the confusion matrix of the selected model (logistic regression, with marginally higher ROC-AUC). The model was chosen on the held-out ROC-AUC of each seed, so its scores are mildly optimistic; the choice is immaterial (0.952 against 0.947), and accuracy, precision, and recall should not be compared across the two models. Both models reached accuracy near 90\% against an adversary designed to imitate the agentic signature, and recall above 96\%, so genuine agent flows were rarely denied priority. The false-accept rate was not small: on the held-out split, 31 of 86 adversarial flows (36.1\%) were accepted as genuine, and applying the trained model to the whole declared-agentic population, which drives the router simulation, accepted 104 of 287 (36.2\%). This leakage is why VATC does not rely on the classifier alone: the rate cap bounds the effect of the flows that leak.

\Table[t!](width=\textwidth){VERIFICATION CLASSIFIER PERFORMANCE, HELD-OUT TEST SET OF 352 FLOWS (SEED 42)\label{tab:classifier}}{\setlength{\NW}{2.0cm}\setlength{\tabcolsep}{4pt}\begin{tabular}{|l|N|N|N|N|N|}
\hline
\textbf{Model} & \textbf{Accuracy} & \textbf{Precision} & \textbf{Recall} & \textbf{F1} & \textbf{ROC-AUC} \\
\hline
Logistic regression & 89.2\% & 89.3\% & 97.4\% & 0.932 & 0.952 \\
Random forest (depth 6, 200 trees) & 89.8\% & 90.5\% & 96.6\% & 0.935 & 0.947 \\
\hline
\end{tabular}}

\Figure[t!](topskip=0pt, botskip=0pt, midskip=0pt)[width=2.6in]{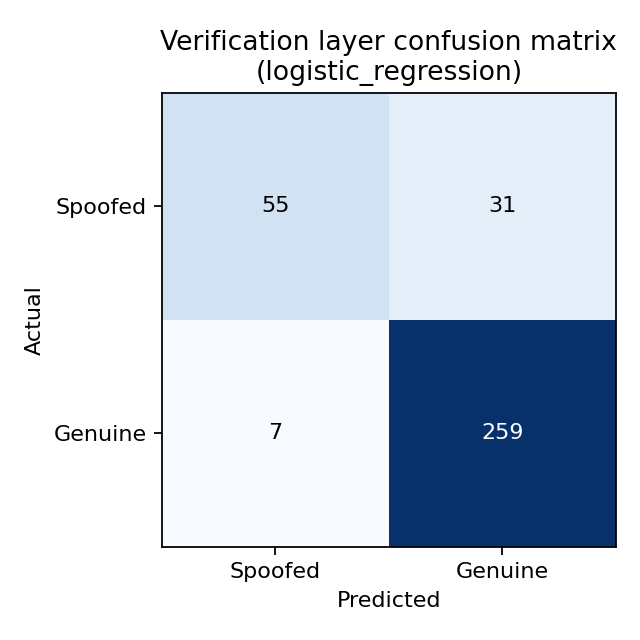}
{Confusion matrix of the logistic-regression verifier on the held-out split (seed 42).\label{fig:confusion}}

\textbf{Accuracy on partial flows. }A router must decide before a flow ends, so the five features were recomputed from the first $n$ packets or first $W$ ticks of each flow and both models retrained, again over 15 seeds (Table~\ref{tab:early}). With the hop-count record available, ROC-AUC fell from 0.955 for complete flows to 0.937 after three packets and 0.948 after ten, and the share of spoofed flows accepted rose from 0.30 to 0.38 and 0.33. Without the hop count, AUC fell to 0.875 for complete flows and 0.79 to 0.85 for partial flows, and leakage was 0.46 to 0.56, so most of the discriminating power comes from the hop-count feature. The hop count is a per-flow attribute in the generator and 92\% of spoofed flows have a hop count of one by construction, so this dependence is partly a property of the generator. 

Windows below about three ticks leave most legitimate flows with fewer than three packets (86\% at two ticks), for which IAT-CV is undefined, so the smallest stable decision point is three packets; accuracy is roughly flat from 3 to 10 packets, and packet windows work better than short time windows (leakage 0.47 to 0.50 at one or two ticks against 0.38 at three packets). This analysis used packet times drawn as in the simulator and not rounded to ticks, so its complete-flow row (leakage 0.30) differs slightly from the main pipeline's and values should be compared only within Table~\ref{tab:early}.

\Table[t!](width=\textwidth){CLASSIFIER ACCURACY ON PARTIAL FLOWS (MEAN OF 15 SEEDS). 5 FEAT. = ALL FIVE FEATURES; 4 FEAT. = WITHOUT THE HOP COUNT; LEAK = FRACTION OF SPOOFED FLOWS ACCEPTED; DEMOTED = FRACTION OF LEGITIMATE FLOWS REJECTED\label{tab:early}}{\setlength{\NW}{2.0cm}\setlength{\tabcolsep}{4pt}\begin{tabular}{|l|N|N|N|N|N|}
\hline
\textbf{Observation window} & \textbf{AUC, 5 feat.} & \textbf{AUC, 4 feat.} & \textbf{Leak, 5 feat.} & \textbf{Leak, 4 feat.} & \textbf{Demoted, 5 feat.} \\
\hline
Complete flow & 0.955 & 0.875 & 0.30 & 0.46 & 0.036 \\
First 3 packets & 0.937 & 0.822 & 0.38 & 0.54 & 0.041 \\
First 5 packets & 0.942 & 0.834 & 0.35 & 0.52 & 0.040 \\
First 10 packets & 0.948 & 0.854 & 0.33 & 0.51 & 0.037 \\
First 1 tick & 0.931 & 0.792 & 0.50 & 0.56 & 0.013 \\
First 2 ticks & 0.936 & 0.804 & 0.47 & 0.54 & 0.012 \\
First 5 ticks & 0.939 & 0.823 & 0.39 & 0.52 & 0.029 \\
\hline
\end{tabular}}

\subsection{Attribution of the latency gain}

The ablation compares scheduling rules and rate constraints one factor at a time, on identical arrival sequences, verification outcomes, packet populations, and FIFO ordering within each class (Section~V-F gives the allocation rule of each variant). Because a weight of 3 against capped strict priority would change both the rate constraint and the scheduling rule, the design crosses the two factors: strict priority and weight 3, each capped and uncapped. All variants use the work-conserving allocator of Algorithm~1, in which capacity a short queue cannot use is redistributed.

Table~\ref{tab:attrib} reports mean agentic and human-surge latency for flood intensities $m = 0$, 1, and 4 (the full sweep covers $m = 0$, 0.5, 1, 2, and 4). Six findings follow.

\Table[t!](width=\textwidth){ATTRIBUTION OF THE GAIN: MEAN LATENCY (TICKS) OF LEGITIMATE AGENTIC (AG.) AND HUMAN-SURGE (HUM.) PACKETS IN THE BURST WITH A SPOOFED FLOOD OF $m \times R$ PACKETS PER TICK, MEAN OF 15 SEEDS\label{tab:attrib}}{\setlength{\NW}{1.7cm}\setlength{\tabcolsep}{4pt}\begin{tabular}{|l|N|N|N|N|N|N|}
\hline
\textbf{Policy} & \textbf{Ag. $m{=}0$} & \textbf{Ag. $m{=}1$} & \textbf{Ag. $m{=}4$} & \textbf{Hum. $m{=}0$} & \textbf{Hum. $m{=}1$} & \textbf{Hum. $m{=}4$} \\
\hline
Best-effort (single queue) & 257.3 & 257.6 & 258.5 & 252.9 & 253.1 & 253.7 \\
Static DiffServ (agentic in BE class) & 23.4 & 23.8 & 24.9 & 18.5 & 18.8 & 19.6 \\
Per-flow fair queuing (no classes) & 36.9 & 38.8 & 38.5 & 27.8 & 27.8 & 27.8 \\
Dedicated class, marking trusted, uncapped & 0.00 & 0.00 & 0.00 & 21.0 & 24.7 & 38.4 \\
Dedicated class, marking trusted, capped & 1.30 & 2.41 & 6.91 & 20.9 & 22.7 & 22.8 \\
Verified, strict priority, uncapped & 0.03 & 0.06 & 0.42 & 20.8 & 24.3 & 36.8 \\
Verified, strict priority, capped (VATC) & 0.46 & 1.48 & 5.48 & 20.8 & 23.8 & 25.4 \\
Verified, weight 3, uncapped & 0.03 & 0.06 & 0.49 & 20.8 & 24.3 & 35.7 \\
Verified, weight 3, capped & 0.46 & 1.48 & 5.48 & 20.8 & 23.8 & 25.4 \\
Verified, weight 1, capped & 0.46 & 1.48 & 5.48 & 20.8 & 23.8 & 25.4 \\
\hline
\end{tabular}}

\emph{First, most of the latency gain over DiffServ comes from giving agent traffic a class of its own}, not from the agent-specific components (Table~\ref{tab:gain}). A class that trusts the marking reduces mean agentic latency from 23.4 ticks (DiffServ) to 0.00 (uncapped) or 1.30 (capped) at $m = 0$, which is 100\% and 94\% of the gain; this is a DiffServ configuration tuned for agents. Verification adds 3.6 to 5.7 percentage points on top of the capped class, so VATC's margin over that configuration is the cap and the verification, not the class.

\Table[t!](width=\textwidth){GAIN ACCOUNTING RELATIVE TO STATIC DIFFSERV, MEAN OF 15 SEEDS, FOR A SPOOFED FLOOD OF $m \times R$ PACKETS PER TICK. AG. = REDUCTION IN THE MEAN LATENCY OF LEGITIMATE AGENTIC PACKETS (\%); HUM. = CHANGE IN THE MEAN LATENCY OF THE CONCURRENT HUMAN SURGE (\%). THE WEIGHT-1 CLASS HAS THE SAME WEIGHT AS BEST-EFFORT IN THE WEIGHTED MAX-MIN ALLOCATOR\label{tab:gain}}{\setlength{\NW}{1.2cm}\setlength{\tabcolsep}{4pt}\begin{tabular}{|l|N|N|N|N|N|N|}
\hline
\textbf{Policy} & \textbf{Ag. $m{=}0$} & \textbf{Ag. $m{=}1$} & \textbf{Ag. $m{=}4$} & \textbf{Hum. $m{=}0$} & \textbf{Hum. $m{=}1$} & \textbf{Hum. $m{=}4$} \\
\hline
Dedicated class, weight 1 (no priority), marking trusted, uncapped & 100 & 99 & 93 & $+$14 & $+$31 & $+$63 \\
Dedicated class, strict priority, marking trusted, uncapped & 100 & 100 & 100 & $+$14 & $+$32 & $+$96 \\
Dedicated class, marking trusted, capped & 94 & 90 & 72 & $+$13 & $+$21 & $+$16 \\
Verified, capped (VATC) & 98 & 94 & 78 & $+$12 & $+$27 & $+$29 \\
Verified, strict priority, uncapped & 100 & 100 & 98 & $+$12 & $+$29 & $+$87 \\
\hline
\end{tabular}}

\emph{Second, isolation, not priority, delivers nearly all of that gain.} A dedicated class with the same weight as best-effort and no priority removed 100\%, 99\%, and 93\% of DiffServ's agentic latency at $m = 0$, 1, and 4, against 100\% for strict priority (Table~\ref{tab:gain}). The class's demand stays below its weighted share of the link, so isolation alone serves it almost without delay. Under the heaviest flood strict priority adds up to 7 points. It also costs the human surge more than a weight of 1 does (96\% against 63\% above DiffServ at $m = 4$).

\emph{Third, the measured value of the cap is that it bounds the collateral cost to human traffic under a flood.} As the flood grows to four times the cap, the human-surge mean rises from 21.0 to 38.4 ticks without a cap (marking trusted) and only to 22.8 (marking trusted) or 25.4 (verified) with one. The capped design had lower human latency than the uncapped design in 15 of 15 seeds at every flood level. The cost is higher agentic latency under a flood (0.46 to 5.48 ticks capped against 0.03 to 0.42 uncapped), still far below the 25 ticks of DiffServ.

\emph{Fourth, the scheduling rule had no measurable effect once the class was capped.} Strict priority, a weight of 3, and a weight of 1 gave identical agentic and human latency when capped (maximum difference 0.0 ticks), and strict priority and a weight of 3 differed by at most 1.1 ticks of human latency when uncapped. The cap, not the rule, is the operative factor.

\emph{Fifth, verification's contribution in this simulator is small}, because the flood is assumed to evade verification and so does not test it. Compared with a capped class that trusts the marking, verification lowered agentic latency slightly, left human latency unchanged without a flood, and raised it by 2.6 ticks at $m = 4$, because demoted flows join a PENALTY class that competes with best-effort (Table~\ref{tab:attrib}). The stronger evidence for verification is the real-kernel ablation of Section~VI-I.

\emph{Sixth, the priority class carries a cost.} Relative to DiffServ at $m = 4$, human-surge latency was 29\% higher with the cap and 87\% higher without it.

\emph{Per-flow fair queuing} did not remove the need for a dedicated class. It was worse than DiffServ for agentic and human traffic (36.9 against 23.4 ticks, and 27.8 against 18.5, at $m = 0$), better for bulk traffic (185.4 against 224.3), and insensitive to the flood: about 2,000 flows start together in the burst, so each receives under one packet per tick, and fairness gives agents and humans equal shares and no priority. This simulator models only the fair-queuing half of FQ-CoDel, without drops, sender feedback, or sparse-flow priority; Section~VI-J reports the real \texttt{fq\_codel}.

\subsection{Robustness across random seeds}

Table~\ref{tab:seeds} reports means and 95\% confidence intervals over 15 seeds for the headline metrics of Sections~VI-A and VI-B. VATC had lower mean agentic latency than DiffServ, and higher human-surge latency, in 15 of 15 seeds. Relative to DiffServ, mean agentic latency fell by 98.0\%, peak latency by 95.3\% (37.2 to 1.7 ticks), and drain delay by 85.6\% (41.7 to 6.0 ticks). The human surge paid 12.2\% more mean latency, 9.1\% more peak latency, and 7.9\% more drain delay. Across the same seeds the logistic-regression classifier reached an accuracy of 89.2\% $\pm$ 0.8\% and a ROC-AUC of 0.954 $\pm$ 0.005. Its leakage was 30.6\% $\pm$ 3.5\% over the whole declared-agentic population and 34.5\% $\pm$ 3.9\% over each seed's held-out adversarial flows (mean 90.7 per seed). The held-out figure is the appropriate estimate, because the population-wide figure includes training data and understates real leakage by several points.

\Table[t!](width=\textwidth){MULTI-SEED ROBUSTNESS: MEAN $\pm$ 95\% CONFIDENCE INTERVAL OVER 15 SEEDS\label{tab:seeds}}{\setlength{\NW}{3.3cm}\setlength{\tabcolsep}{4pt}\begin{tabular}{|l|N|N|N|}
\hline
\textbf{Metric} & \textbf{Best-effort} & \textbf{Static DiffServ} & \textbf{VATC} \\
\hline
Steady-state human mean latency (ticks) & 182.137 $\pm$ 4.042 & 0.001 $\pm$ 0.001 & 0.000 $\pm$ 0.000 \\
Steady-state bulk completion rate & 78.2\% $\pm$ 0.8\% & 71.4\% $\pm$ 1.0\% & 71.4\% $\pm$ 1.0\% \\
Burst agentic mean latency (ticks) & 257.26 $\pm$ 10.07 & 23.36 $\pm$ 0.29 & 0.47 $\pm$ 0.05 \\
Burst human-surge mean latency (ticks) & 252.87 $\pm$ 10.08 & 18.52 $\pm$ 0.16 & 20.78 $\pm$ 0.25 \\
\hline
\end{tabular}}

\subsection{Cross-check against socket-measured traffic}

To check whether the assumed direction of the differences appears in measured sessions, three client and server archetypes (agentic, human, bulk) were run over TCP loopback sockets, 60 sessions each, and the four network-observable features were computed when each socket call completed (Table~\supref{tab:socket} of the Supplementary Material). Agentic sessions ran longer than human bursts (mean 0.50 against 0.25~s; Kolmogorov--Smirnov statistic KS = 0.53) and than bulk transfers, which finished in milliseconds on loopback (KS = 1.00), and their inter-arrival timing was smoother than that of human and bulk traffic (KS = 1.00 for both). A minority of agentic sessions (6.7\% of 60) fell in the heavy-upstream tail against 0.0\% of human and bulk sessions, consistent in direction with the 9\% against 0.5\% reported by Cisco. The check is on direction and ordering only; it is a sandbox-scale test, and because the archetypes were designed by the authors, agreement is partly by construction.

\subsection{Second-simulator consistency check}

The router experiment was reimplemented with an event-driven arrival process, packet-level Deficit Round Robin (DRR) scheduling \cite{drr}, and an independent token-bucket policer for the AGENTIC class. Both engines ran on the identical flow population and verification outcomes, so the check covers the scheduler code only (Tables~\supref{tab:drrsteady} and~\supref{tab:drrburst} of the Supplementary Material). Best-effort rows agree to within 4\% (agentic 181.0 against 180.7 ticks in steady state; 248.8 against 240.1 in the burst), and so does the bulk latency under the differentiated policies (266.0 against 266.4 ticks). Other absolute values differ, as expected from DRR's discrete quantum, but the engines agree on every qualitative conclusion: near-zero human and agentic latency under both differentiated policies in steady state, and, in the burst, VATC reducing agentic mean latency well below DiffServ (0.5 ticks under both engines) at a small cost to the human surge (22.0 against 21.3 ticks). This is not validation on ns-3 or hardware.

\subsection{Real-kernel validation under honest load}

The consistency check of Section~VI-G isolates the scheduling algorithm but still runs inside a Python model. To test the mechanism on a real network stack, the two harnesses of Section~V-H were used with VATC's actual scheduling configuration, Linux HTB with strict-priority classes and DSCP-based classification, on real TCP traffic. All policies (best-effort, static DiffServ, VATC) and both phases (steady and burst) were run end to end in both harnesses. All 4,320 TCP sessions completed without failure, and kernel class counters (\texttt{tc -s -d class show}) showed no packet drops on the AGENTIC class. Table~\ref{tab:kernelburst} reports the burst mean latency in milliseconds; the steady-state latencies, which did not separate the differentiated policies, are in Table~\supref{tab:kernelsteady} of the Supplementary Material. In the netns harness the best-effort configuration is an unshaped queue on the virtual link and is not congested, so the netns best-effort rows are not a valid comparison with the shaped policies; Mininet is the reference for best-effort.

\Table[t!](width=\textwidth){REAL-KERNEL BURST MEAN LATENCY (MS), HONEST LOAD\label{tab:kernelburst}}{\setlength{\NW}{2.0cm}\setlength{\tabcolsep}{4pt}\begin{tabular}{|l|N|N|N|N|}
\hline
\textbf{Policy} & \textbf{Ag., Mininet} & \textbf{Ag., netns} & \textbf{Hum., Mininet} & \textbf{Hum., netns} \\
\hline
Best-effort & 41.55 & 0.029 & 56.04 & 0.059 \\
Static DiffServ & 4.22 & 0.039 & 4.17 & 0.057 \\
VATC & 4.33 & 0.085 & 5.00 & 0.046 \\
\hline
\end{tabular}}

In the burst, both differentiated policies cut agentic mean latency in Mininet from 41.55~ms (best-effort) to 4.22~ms (DiffServ, 89.8\% lower) and 4.33~ms (VATC, 89.6\% lower), and human latency from 56.04~ms to 4.17 and 5.00~ms. Unlike the two simulators, the real-kernel test did not separate VATC from DiffServ under honest load: 4.33 against 4.22~ms in Mininet, and 0.085 against 0.039~ms in netns, both at the harness's noise floor. The likely reason is that this experiment contains only well-behaved load: at about 40~Mbit/s of genuine agentic traffic on a 200~Mbit/s link, the borrowing of DiffServ's best-effort class already gives agentic traffic latency comparable to VATC's non-borrowing 60~Mbit/s cap. VATC's design advantage is a guarantee against a misbehaving flow, which this experiment did not exercise. HTB rate-limited 8.7\% (Mininet) and 11.4\% (netns) of AGENTIC packets with no drops, consistent with a strict token bucket pacing individual packets against micro-bursts. Bulk latency in the burst rose to 97.6--105.1~ms under VATC and DiffServ, the intended cost of the scavenger class. The test therefore supports the gain over best-effort on a real scheduler and real TCP, and separating VATC from a simpler static-priority baseline requires the spoofed-flow experiments below.

\subsection{Real-kernel evaluation under spoofed traffic}

Both harnesses were extended with a spoofed source: a bulk-shaped continuous transfer declared with the AGENTIC marking, representing traffic that leaked past verification (Section~V-H). The swarm was 20 sessions, injected in the burst alongside the 180 honest agentic sessions with verification off, so that all 20 were admitted to the AGENTIC class. A refinement was tested alongside plain VATC: VATC-fair is identical except that its AGENTIC leaf queue is per-flow SFQ instead of a single FIFO. VATC-unbounded is the real-kernel analogue of the uncapped design. Table~\ref{tab:spoof} reports burst mean latency under this load.

\Table[t!](width=\textwidth){REAL-KERNEL BURST MEAN LATENCY (MS) WITH 20 SPOOFED SESSIONS INJECTED ALONGSIDE 180 LEGITIMATE AGENTIC SESSIONS (SEED 42)\label{tab:spoof}}{\setlength{\NW}{3.2cm}\setlength{\tabcolsep}{4pt}\begin{tabular}{|l|N|N|N|N|}
\hline
\textbf{Policy} & \textbf{Legit. agentic, Mininet} & \textbf{Legit. agentic, netns} & \textbf{Spoofed, Mininet} & \textbf{Spoofed, netns} \\
\hline
Best-effort & 49.62 & 0.032 & 58.22 & 0.035 \\
Static DiffServ & 3.16 & 0.60 & 4.35 & 1.67 \\
VATC & 29.23 & 21.25 & 30.65 & 27.59 \\
VATC-fair (SFQ leaf) & 13.15 & 13.84 & 26.31 & 24.79 \\
VATC-unbounded & 2.25 & 0.45 & 3.07 & 0.99 \\
\hline
\end{tabular}}

Legitimate agentic latency under plain VATC was much higher than under DiffServ in both harnesses. Although the spoofed flows were only 20 of 200 AGENTIC-marked sessions, they were bulk-shaped and carried about 30 times more bytes per session than legitimate sessions (5.4 to 5.7 against 0.18 to 0.19~Mbit/s). Combined AGENTIC demand reached about 150~Mbit/s against the 60~Mbit/s cap, which had been sized for legitimate load alone, and HTB rate-limited 46\% to 48\% of AGENTIC packets with no drops. Latency shifted smoothly across sessions, consistent with aggregate self-congestion in the class and not with a few starved flows. The cap protected the other classes: human latency (2.29 to 3.97~ms) and bulk latency (117.4 to 123.9~ms) differed little across DiffServ, VATC, VATC-fair, and VATC-unbounded, and below best-effort. The cap did not, however, protect one legitimate flow from a spoofed flow sharing its class and queue; that separation is the job of verification.

VATC-fair roughly halved legitimate agentic latency relative to plain VATC (29.23 to 13.15~ms in Mininet; 21.25 to 13.84~ms in netns) with no change to human or bulk latency, although a few sessions had a longer tail (3 of 180 in Mininet and 7 of 180 in netns), as expected when hash-bucketed fair queuing lets a flow collide with a greedy one. VATC-unbounded gave the lowest legitimate agentic latency at this swarm size because the combined demand of about 150~Mbit/s saturated the cap but not the link. That is not evidence that the uncapped design is preferable: the recommended ceiling for this swarm (224~Mbit/s) already exceeds the 200~Mbit/s link, so no ceiling below link rate could absorb this flood without queuing, and a larger flood would let an uncapped class take capacity from human and bulk traffic, as in the simulator ablation of Section~VI-D.

\textbf{Cap ceiling. }Raising the AGENTIC ceiling from 60 to 180~Mbit/s lowered legitimate agentic latency at every step, from 20.6 to 0.45~ms under plain VATC (Table~\ref{tab:ceiling}; five seeds, 20-session flood, verification off), and the ranges over the five seeds did not overlap between adjacent ceilings. Human latency stayed between 0.05 and 0.08~ms up to 150~Mbit/s and rose to 0.19~ms at 180~Mbit/s, still below DiffServ's 0.65~ms. The gain came from admitting the flood, not from stopping it: spoofed sessions also gained, from 29 to 1.1~ms, because the lane was no longer saturated. The cost to human traffic is therefore confined to the top of the range tested, between 150 and 180~Mbit/s.

\Table[t!](width=\textwidth){EFFECT OF THE AGENTIC CEILING ON BURST MEAN LATENCY (MS) UNDER THE 20-SESSION SPOOFED FLOOD, VERIFICATION OFF, NETNS HARNESS, MEAN $\pm$ STANDARD DEVIATION OVER FIVE SEEDS (42 TO 46). AG. = LEGITIMATE AGENTIC; HUM. = HUMAN. DIFFSERV GAVE 0.62 AND 0.65~ms (TABLE~\ref{tab:fqcodel})\label{tab:ceiling}}{\setlength{\NW}{2.5cm}\setlength{\tabcolsep}{4pt}\begin{tabular}{|N|N|N|N|N|}
\hline
 & \multicolumn{2}{c|}{\textbf{VATC}} & \multicolumn{2}{c|}{\textbf{VATC-fair}} \\
\hline
\textbf{Ceiling (Mbit/s)} & \textbf{Ag.} & \textbf{Hum.} & \textbf{Ag.} & \textbf{Hum.} \\
\hline
60 & 20.6 $\pm$ 2.6 & 0.070 $\pm$ 0.031 & 11.4 $\pm$ 1.4 & 0.048 $\pm$ 0.004 \\
90 & 8.0 $\pm$ 0.8 & 0.052 $\pm$ 0.007 & 4.0 $\pm$ 1.0 & 0.045 $\pm$ 0.005 \\
120 & 1.93 $\pm$ 0.17 & 0.057 $\pm$ 0.006 & 1.38 $\pm$ 0.23 & 0.056 $\pm$ 0.005 \\
150 & 0.80 $\pm$ 0.07 & 0.077 $\pm$ 0.039 & 0.66 $\pm$ 0.14 & 0.063 $\pm$ 0.012 \\
180 & 0.45 $\pm$ 0.04 & 0.187 $\pm$ 0.068 & 0.35 $\pm$ 0.05 & 0.168 $\pm$ 0.048 \\
\hline
\end{tabular}}

\textbf{Swarm size and verification. }The netns experiment was repeated at four swarm sizes (10 to 40 spoofed sessions with 180 legitimate sessions, three seeded trials per size; Table~\supref{tab:swarm} of the Supplementary Material). With verification on, each spoofed session was assigned at generation time to ``caught'' (demoted to PENALTY) or ``missed'' using the held-out miss rate of 34.5\%, a statistical proxy and not a live run. The proxy removed about 65\% of spoofed sessions from the AGENTIC class and lowered legitimate agentic latency by a factor of 4.0 to 12.7 for plain VATC and 2.5 to 6.7 for VATC-fair; the factor shrank as the swarm grew because verification bounds the fraction of adversarial traffic that reaches the capped class, not its absolute number. Without verification, VATC-fair reduced latency relative to plain VATC by 29\% to 63\%, and by 7\% to 41\% with it.

\textbf{Live verifier. }Replacing the statistical proxy with a classifier running at the shaped interface (Section~V-H, Table~\ref{tab:live}) removed most of the flood's effect on legitimate agents. Under the 20-session flood with the cap at 60~Mbit/s, it demoted all 20 spoofed sessions in each of five seeds, a median 36~ms (95th percentile 52~ms) after the flow's first data packet, and demoted none of the 900 legitimate burst sessions or the 300 steady-state sessions. Legitimate agentic latency fell from 20.1 to 0.30~ms, below DiffServ (0.62~ms) and tight \texttt{fq\_codel} (3.80~ms; Table~\ref{tab:fqcodel}), and human latency was 0.056~ms against 0.65~ms for DiffServ. The demoted flows paid for it: their latency was 125~ms in the PENALTY class against 29~ms in the AGENTIC lane.

This separation is cleaner than the 35\% leakage of the proxy implies, because the two measured different adversaries. The kernel's spoofers were bulk-shaped flows that moved about 30 times as many bytes per session as an agent session, and duration and throughput alone separated them in the training data (training accuracy 1.000). The 34.5\% miss rate was measured in the simulator on adversarial flows that overlap the agentic profile. To approach that case, six mimic sessions were added that copied the agent message pattern with larger payloads, declared AGENTIC. The live verifier demoted none of them in any seed (0 of 30). Their latency was 0.42~ms, comparable to that of legitimate agents (0.53~ms), because six sessions did not saturate the cap; what bounded the harm of a mimic was the cap, not the verifier. A model trained with the mimics as negative examples and scored offline on the same flows would have demoted 33\% of them and 1\% of legitimate flows, so these four features do not separate the mimics reliably either. With the mimics present, legitimate agentic latency was 0.53~ms, and two of the 900 legitimate burst sessions (0.2\%) were demoted, both with an upstream byte share of 0.92 to 0.94, in the range of the upload-heavy sessions that the generator includes. Human latency averaged 0.32~ms because one seed gave 1.36~ms; the other four gave 0.04 to 0.07~ms.

\Table[t!](width=\textwidth){LIVE INLINE VERIFIER IN THE NETNS HARNESS: BURST MEAN LATENCY (MS) OF LEGITIMATE AGENTIC AND HUMAN SESSIONS (MEAN $\pm$ STANDARD DEVIATION OVER FIVE SEEDS, 42 TO 46), SESSIONS DEMOTED TO PENALTY (POOLED OVER SEEDS), AND MEDIAN TIME FROM A FLOW'S FIRST DATA PACKET TO ITS DEMOTION. EVERY RUN HAS 20 SPOOFED BULK-SHAPED SESSIONS; MIMICS ARE SIX SPOOFED SESSIONS THAT COPY THE AGENT MESSAGE PATTERN. NO VERIFIER IS VATC WITH VERIFICATION OFF\label{tab:live}}{\setlength{\NW}{1.85cm}\setlength{\tabcolsep}{4pt}\begin{tabular}{|l|N|N|N|N|N|N|N|}
\hline
\textbf{Verifier} & \textbf{Mimics} & \textbf{Agentic} & \textbf{Human} & \textbf{Spoofed demoted} & \textbf{Mimics demoted} & \textbf{Legit. demoted} & \textbf{Verdict (ms)} \\
\hline
None & 0 & 20.09 $\pm$ 3.04 & 0.050 $\pm$ 0.004 & -- & -- & -- & -- \\
Live & 0 & 0.30 $\pm$ 0.12 & 0.056 $\pm$ 0.018 & 100 of 100 & -- & 0 of 900 & 36 \\
None & 6 & 21.67 $\pm$ 4.03 & 0.100 $\pm$ 0.104 & -- & -- & -- & -- \\
Live & 6 & 0.53 $\pm$ 0.28 & 0.317 $\pm$ 0.585 & 100 of 100 & 0 of 30 & 2 of 900 & 37 \\
\hline
\end{tabular}}

Taken together, the three layers are complementary and were quantified separately: verification reduces how much adversarial traffic is admitted (for bulk-shaped spoofers; a mimic passes), the cap bounds what the admitted portion can take in aggregate, and the SFQ leaf protects individual legitimate flows from adversarial traffic that still shares their queue. The artifact repository \cite{artifact} contains the per-session data, kernel counters, and sizing-check methodology.

\subsection{Comparison with fq\_codel on the real kernel}

The attribution study (Section~VI-D) modeled only per-flow fair queuing. The real Linux \texttt{fq\_codel} scheduler (Section~V-G) was therefore run in the netns harness over five seeds (42 to 46), with and without the 20-session spoofed flood, and in the Mininet harness at one seed without a flood. Kernel counters confirmed that it was active (1,892 packets dropped in netns and 2,173 in Mininet; no explicit congestion notification marks, because the traffic does not use it). Table~\ref{tab:fqcodel} reports the netns results.

\Table[t!](width=\textwidth){REAL-KERNEL (NETNS) BURST MEAN LATENCY (MS) OF DIFFSERV, VATC, AND fq\_codel AT THREE PARAMETER SETTINGS, MEAN $\pm$ STANDARD DEVIATION OVER FIVE SEEDS (42 TO 46), ONE RUN PER SEED. AG. = AGENTIC; HUM. = HUMAN; SPOOF. = SPOOFED. THE FLOOD IS THE 20-SESSION SPOOFED SWARM\label{tab:fqcodel}}{\setlength{\NW}{2.1cm}\setlength{\tabcolsep}{4pt}\begin{tabular}{|l|N|N|N|N|N|}
\hline
 & \multicolumn{2}{c|}{\textbf{No flood}} & \multicolumn{3}{c|}{\textbf{Flood}} \\
\hline
\textbf{Policy} & \textbf{Ag.} & \textbf{Hum.} & \textbf{Ag.} & \textbf{Hum.} & \textbf{Spoof.} \\
\hline
Static DiffServ & 0.047 $\pm$ 0.004 & 0.058 $\pm$ 0.003 & 0.62 $\pm$ 0.10 & 0.65 $\pm$ 0.13 & 2.31 $\pm$ 0.20 \\
VATC & 0.11 $\pm$ 0.01 & 0.054 $\pm$ 0.009 & 19.9 $\pm$ 2.0 & 0.059 $\pm$ 0.009 & 29.2 $\pm$ 2.9 \\
fq\_codel, default (5~ms, 100~ms) & 11.1 $\pm$ 1.4 & 10.2 $\pm$ 1.4 & 13.3 $\pm$ 1.4 & 10.8 $\pm$ 1.3 & 120.6 $\pm$ 6.0 \\
fq\_codel, tuned (1~ms, 20~ms) & 6.40 $\pm$ 1.31 & 4.99 $\pm$ 0.70 & 5.99 $\pm$ 1.30 & 5.58 $\pm$ 1.52 & 118.0 $\pm$ 2.7 \\
fq\_codel, tight (200~$\mu$s, 5~ms) & 3.72 $\pm$ 0.82 & 1.73 $\pm$ 0.91 & 3.80 $\pm$ 0.71 & 2.33 $\pm$ 0.79 & 122.3 $\pm$ 7.5 \\
\hline
\end{tabular}}

\emph{Without a flood}, \texttt{fq\_codel} at every setting was slower than both class-based policies for agentic and human traffic (Table~\ref{tab:fqcodel}), and the ranges over the five seeds did not overlap with those of VATC (agentic: 9.3 to 12.6~ms for the default setting, 2.6 to 4.6~ms for the tight setting, 0.09 to 0.12~ms for VATC). Tuning helped monotonically, but even the tight setting was about 30 to 35 times slower than VATC, although the absolute values were small. In Mininet at one seed the gap was smaller: the tight setting gave 6.12 and 7.75~ms against 4.64 and 4.72~ms for VATC, 3.83 and 4.61~ms for DiffServ, and 20.7 and 26.3~ms for the default setting. Single Mininet runs vary by several milliseconds, so the Mininet figures are indicative only.

\emph{Under the flood}, the ordering of legitimate agentic latency changed. DiffServ and all three \texttt{fq\_codel} settings gave legitimate agents lower latency than VATC (Table~\ref{tab:fqcodel}; for example 0.62~ms for DiffServ, 3.80~ms for the tight setting, and 19.9~ms for VATC). The tight setting was better than VATC in each of the five seeds (2.9 to 4.7~ms against 17.8 to 23.1~ms), because VATC's lane was shared with a spoofed swarm whose demand of about 150~Mbit/s exceeded the 60~Mbit/s cap (Section~VI-I). With the live verifier of that section, which demoted the spoofed sessions, VATC gave 0.30~ms. Human traffic fared differently: VATC held human latency at 0.059~ms against 0.65~ms for DiffServ, and the tight setting was about 40 times slower than VATC for humans (2.33~ms). Spoofed flows were delayed by about 120~ms under every \texttt{fq\_codel} setting, against 29~ms under VATC and 2.3~ms under DiffServ, because per-flow fair queuing slows heavy flows whatever their marking.

\emph{Does \texttt{fq\_codel} remove the need for a dedicated class?} In part. Tuned, it protected legitimate agent traffic under a spoofed flood better than VATC with a cap sized for legitimate load, without any marking, verification, or class configuration, and in the real-harness measurement of Section~VI-L it recovered nearly as much of the task time as DiffServ and VATC (96\% to 98\%, against 97\% to 100\% for DiffServ). It did not remove the queuing gap in the correlated honest burst, where the simulated per-flow fair queuing was slower than DiffServ for agentic and human traffic (Table~\ref{tab:attrib}), and the default kernel setting was far slower than either class-based policy. What these results do not show is that no \texttt{fq\_codel} setting can match VATC, because the improvement was still monotone at the tightest setting tested, and only three settings were tried without a search for an optimum. The cost of \texttt{fq\_codel} in these tests was higher latency for human traffic than a policed class gave, and no bound on what a spoofed flow can take other than fairness among flows. Combining per-flow fair queuing within a class, which the VATC-fair variant of Section~VI-I approximates, therefore looks more promising than choosing between the two.

\subsection{Probation of new flows}

Probation (Section~IV-D) was evaluated in the simulator over 15 seeds; the full text and Tables~\supref{tab:probnoattack} to \supref{tab:vdelay} are in the Supplementary Material. The 400 legitimate burst flows started together (ramp 0) or over 20 ticks (ramp 20), and a flow declared AGENTIC was probationary for its first $N$ packets. Four treatments were compared: the verdict known at the first packet (oracle), full priority (optimistic), best-effort, and the sub-budget lane with $R_p$ equal to 20\% of $R$.

Without an attacker, the sub-budget lane cost little more than optimistic probation (0.6 ticks at $N = 3$ and ramp 20), whereas best-effort probation grew from 5.88 to 29.82 ticks as $N$ went from 3 to 20 (Table~\supref{tab:probnoattack}). Two micro-burst adversaries, declared AGENTIC with a demand of twice the cap, tested priority theft: a fast adversary sending 20-packet flows and a window-matched adversary sending flows of $N - 1$ packets, so that every packet was probationary (Table~\supref{tab:probattack}). Under optimistic probation the window-matched adversary took 65\% to 69\% of priority service, with peaks up to the cap. Under the sub-budget lane the adversaries took 7\% to 14\%, with peaks below $R_p$. At $N = 3$ with the window-matched adversary, legitimate agentic latency was 19.8 and 30.0 ticks (ramp 0 and 20) under optimistic probation against 5.1 and 0.84 under the sub-budget lane (Table~\supref{tab:probearly}). When the verdict was computed from the first $N$ packets, the classifier let 33\% to 37\% of adversarial flows through and demoted about 4\% of legitimate ones. A 20-packet adversary that evaded verification at that rate still took 44\% of priority service under the lane, so the rate cap, not probation, bounds that case.

A delay $D$ between the $N$-th packet and the installed verdict, as with a verifier behind a flow exporter, progressively eroded the gain (Table~\supref{tab:vdelay}). Legitimate agentic latency under the lane rose from 0.76 ticks at $D = 0$ to 23.5 at $D = 20$, against 29.7 under static DiffServ, so the reduction relative to DiffServ fell from 97\% to 21\% (89\% and 68\% at $D$ of 2 and 5). At $D = 20$, 74\% of legitimate agentic flows ended before their verdict. The 98\% reduction of Section~VI-B therefore assumes an immediate verdict and is an upper bound. The results support $N$ of 3 to 5 packets with $R_p$ equal to 20\% of $R$. The lane protects against flows that would be demoted or never receive a verdict, not against flows that pass verification, and the adversaries were fixed designs that did not adapt to $R_p$; the lane was simulated only.

\subsection{Router queuing in the task time of real agent harnesses}

Table~\ref{tab:rh} gives the measured task times. With the single FIFO and the 1,000-packet buffer, the router queue averaged about 185~ms in each direction (95th percentile about 400~ms), and each model call spent about 0.9~s in the network, against 0.09~s without background load. Harness B, which also made remote tool calls and opened an MCP session, was delayed about twice as much as harness A. Because the added delay did not depend on the model, its share of task time fell as the assumed model latency rose between 0.5 and 5~s per call: from 48\% to 13\% for harness A and from 63\% to 24\% for harness B. With the 250-packet buffer, where the queue averaged 40 to 50~ms, the share at 2~s per call was 7\% for harness A and 13\% for harness B.

\Table[t!](width=\textwidth){ROUTER QUEUING IN THE TASK TIME OF TWO REAL AGENT HARNESSES (A: MINI-SWE-AGENT; B: OPENAI-CLIENT AND MCP-SDK LOOP). MODEL = ASSUMED MODEL LATENCY PER CALL (MEDIAN, s); BUFFER = ROUTER FIFO (PACKETS). TASK TIME (s) WITHOUT BACKGROUND LOAD AND WITH A SINGLE FIFO UNDER BULK AND WEB LOAD. QUEUE SHARE = (FIFO $-$ NO LOAD) / FIFO, \% (95\% BOOTSTRAP INTERVAL OVER PAIRED TASKS). DIFFSERV AND VATC = SHARE OF THE EXTRA TASK TIME REMOVED BY STATIC DIFFSERV, BY THE VATC DATA PLANE, AND BY FQ-CODEL AT ITS DEFAULT AND TUNED (1 ms, 20 ms) SETTINGS (\%; -- = NOT RUN). 48 PAIRED TASKS PER ROW (24 FOR THE 250-PACKET ROWS)\label{tab:rh}}{\setlength{\NW}{1.45cm}\setlength{\tabcolsep}{4pt}\begin{tabular}{|l|N|N|N|N|N|N|N|N|N|}
\hline
\textbf{Harness} & \textbf{Model} & \textbf{Buffer} & \textbf{No load} & \textbf{FIFO} & \textbf{Queue share} & \textbf{DiffServ} & \textbf{VATC} & \textbf{FQ-CoDel} & \textbf{FQ-CoDel tuned} \\
\hline
A & 0.5 & 1{,}000 & 6.5 & 12.6 & 48 (47--49) & 99 & 96 & 98 & 97 \\
A & 2 & 1{,}000 & 19.4 & 26.1 & 26 (24--26) & 99 & 98 & 98 & 97 \\
A & 5 & 1{,}000 & 45.3 & 52.0 & 13 (12--13) & 100 & 99 & -- & -- \\
B & 0.5 & 1{,}000 & 7.0 & 18.8 & 63 (62--64) & 97 & 96 & 97 & 96 \\
B & 2 & 1{,}000 & 19.9 & 31.8 & 37 (36--38) & 97 & 98 & 97 & 96 \\
B & 5 & 1{,}000 & 45.9 & 60.0 & 24 (22--24) & 97 & 98 & -- & -- \\
\hline
A & 2 & 250 & 19.5 & 21.1 & 7 (7--8) & 93 & 89 & -- & -- \\
B & 2 & 250 & 20.1 & 23.1 & 13 (12--14) & 90 & 92 & -- & -- \\
\hline
\end{tabular}}

Static DiffServ removed 97\% to 100\% of the extra task time, and the VATC data plane 96\% to 99\%. Demoting the bulk transfers accounted for the improvement. The dedicated AGENTIC class added no measurable benefit in this scenario: VATC was within 0.2~s of DiffServ in both harnesses (0.04 to 0.19~s slower for harness A, which may reflect the 5~Mbit/s cap serializing large prompts but was not tested). Agent traffic and web load were light and did not burst together. Median web page-load time fell from 3.8~s with the single FIFO to 1.1~s with DiffServ and 1.0~s with VATC. Linux \texttt{fq\_codel} at its default setting removed 97\% to 98\% of the extra time and at the tuned setting 96\% to 97\% (0.5 and 2~s per call), as expected of per-flow fair queuing for a few short flows next to bulk transfers. The measurement supports the claim that queuing can take a material share of agent task time on a congested link, and it is consistent with Section~VI-D: class isolation, not the agent-specific class, supplies most of the gain. Light agent and web load does not separate disciplines that isolate the bulk transfers; the advantage of the AGENTIC class appears when agent and human traffic burst together or spoofed traffic is present (Sections~VI-B, VI-D, and VI-J).

\section{Discussion}

\subsection{Deployment path}

VATC is designed for incremental deployment on DiffServ-capable software and edge routers. The AQI-to-DSCP mapping needs no new packet format, only a reserved codepoint pool, following the pattern already specified for 5QI-to-DSCP mapping \cite{cbs}. Verification runs on flow records rather than payloads, and four of its five features are commonly exported; the fifth, the hop count, needs a trusted gateway and the out-of-band record of Section~IV-C, and where that record is absent verification degrades to the other four features, at the cost of the precision shown in Table~\ref{tab:early}. Because verification takes time (Section~IV-A), the rate cap, not the classifier, bounds the harm from a spoofed flow before it is demoted, and the probation lane's benefit shrinks as the verdict is delayed (Section~VI-K). The cap $R$ must be sized from the expected legitimate agent demand and the spoofed traffic that verification is expected to admit: a cap sized for legitimate load alone is shared with whatever leaks past verification. That is why plain VATC without verification was slower than DiffServ for legitimate agents in the kernel flood test, and why raising the ceiling toward the combined demand removed the gap (Section~VI-I). A cap far above legitimate demand, however, weakens the bound on human delay.

\subsection{Security considerations}

The threat model is priority theft: a flow that falsely claims agentic status to obtain preferential forwarding. The complementary threat of an adversary that controls or compromises a genuine agent framework, and so produces traffic with a legitimately attested but maliciously intended marking, is not addressed; it is a trust and identity problem for which the IETF agent identity work \cite{aip} is a natural point of integration.

A related trust boundary concerns the verifier's strongest feature. The other four features are derived from network- and transport-layer observables and require no cooperation from higher layers, whereas the hop count is an application-layer signal that cannot be derived from packet timing, sizes, or headers, and its resistance to spoofing is only as strong as the party that supplies it. The early-flow analysis (Section~VI-C) showed that most of the classifier's discriminating power comes from this feature, so a fabricated hop count would remove most of the benefit of verification. The full benefit therefore depends on an independent gateway that observes and attaches the tool-call chain, and the safe default without one is the four-feature fallback.

A further question is why the router should verify behavior at all if a trusted gateway supplies the hop count: the gateway could sign or attest the marking, and routers would give priority only to traffic carrying a valid token. That alternative admits no unauthenticated spoofer but needs an in-band carrier, (with the difficulties of Section~IV-C) and key distribution to every router, and a signature attests who sent the traffic and not what it does: an authorized agent that sent a bulk transfer under the agentic marking would keep its priority, limited only by the rate cap. Behavioral verification checks the traffic against the declared class and works, through the four-feature fallback, where no trusted gateway exists. The two approaches are complementary, and a quantitative comparison is left for future work.

The evaluated adversaries were fixed designs, apart from the mimic spoofers of Section~VI-I; an adversary that adapts to the probation sub-budget or to the verifier's features was not tested.

\subsection{WAN round-trip times and TCP dynamics}

The kernel tests of Sections~VI-H and VI-I ran on local virtual interfaces with negligible propagation delay and a single RTT, with CUBIC as the congestion-control algorithm. Agent sessions that reach remote inference services see RTTs of tens to over a hundred milliseconds, and their sending rates are shaped by TCP congestion control. Pacing and the bandwidth-probing phases of Bottleneck Bandwidth and Round-trip propagation time (BBR) congestion control change packet spacing \cite{bbr}, and delay changes where the server's packets fall relative to the client's, so the IAT-CV observed at the router can differ from its value on a local link. A real-kernel measurement and a sensitivity analysis address this; the full text and Tables~\supref{tab:wanmeas} to \supref{tab:wansens} are in the Supplementary Material.

In three network namespaces joined by virtual Ethernet pairs, with \texttt{netem} delay on the router-to-server link, the three traffic archetypes ran over real TCP at RTTs of 0, 20, 80, and 150~ms with CUBIC and BBR (30 sessions per archetype and cell, two repetitions). Delay changed the IAT-CV of human and agentic flows more than the congestion-control algorithm did. At 0~ms an agentic flow had a lower IAT-CV than a human flow with probability 0.88 to 0.89 (AUC); at 20 to 80~ms this fell to 0.40 to 0.44, and at 150~ms to 0.22 to 0.27, below chance. IAT-CV still separated agentic flows from bulk transfers (AUC 1.00). A verifier trained on 0~ms flows did not transfer: logistic regression accepted every human and bulk flow at 80 and 150~ms, and a threshold chosen at 0~ms demoted 68\% to 75\% of agentic flows at 150~ms. Training at the same RTT or on pooled RTTs restored a usable operating point at 20~ms and only partly at 80 and 150~ms (3\% to 18\% of agentic flows rejected and 1\% to 24\% of human and bulk flows accepted, against 0\% and 0\% at 0~ms), and adding the injected hop count reduced the errors. These results apply to one feature of three synthetic archetypes on one virtual-machine testbed with exact emulated delay, uniform RTTs within a run, and no loss or jitter.

Two mitigations apply. The verifier should be calibrated or retrained on flows captured on paths representative of the deployment; the four-feature fallback (Section~VII-A) is not a remedy, because the timing features are the ones affected. The rate cap and the probation lane (Sections~IV-D and IV-E) bound the cost of a wrong verdict, since a wrongly demoted flow loses priority and a wrongly accepted flow is limited to its share of the capped class. One caveat remains: a measurement on physical networks and on real agent traffic is still needed before the reported accuracy is assumed to hold on wide-area paths.

\subsection{Limitations}

Several limitations apply. First, the traffic is synthetic and parameterized from the statistics that motivate the design; the generator gives 92\% of spoofed flows a hop count of one, so classifier results partly describe the generator; the socket-measured sessions (Section~VI-F) were designed by the authors and are a directional check only; and no production agent traces were available.

Second, the primary simulator is a custom discrete-time model, checked against a separately implemented simulator (which shares the flow population and verdicts, so it checks the scheduler code only) and against real Linux-kernel queuing. The kernel tests used local interfaces with negligible propagation delay, a single RTT, and CUBIC; the delay measurement (Section~VII-C) showed that IAT-CV stopped separating agentic from human flows and that a verifier trained on local-link flows failed on delayed paths until it was retrained. The ceiling sweep, live-verifier test, and \texttt{fq\_codel} comparison used five seeds, the Mininet results one, and ns-3 or hardware validation at production scale was not done. Under honest load the kernel tests did not separate VATC from DiffServ. The probation lane was evaluated only in the simulator; a kernel implementation would need a per-flow packet counter, for example in an eBPF map, and a second strict-priority class.

Third, the classifier is lightweight and was scored on complete flows; accuracy on partial flows is lower and depends on the hop count (Section~VI-C). The hop-count channel and the export latency of a real IPFIX exporter were not exercised, the verdict delay was simulated (Section~VI-K), and inference and verdict installation were timed only in isolation (Section~IV-A). The verification contribution in the simulator is small because the flood was assumed to evade it. The stronger evidence is a statistical proxy and a live four-feature classifier in the kernel, which has no hop-count feature, leaves flows in the AGENTIC class until the verdict, reads packets at a tap after the egress queue, was trained on the generator that produced its test traffic, and demoted none of six spoofers that copied the agent message pattern.

Fourth, the evaluation used one AQI class and one router hop. Deployment would need a multi-class enumeration, consistent marking across domains, and a fallback to best-effort where a domain does not recognize the reserved codepoint pool, because DSCP bleaching at administrative boundaries is not addressed and RFC~8100 shows that interconnection classes must be agreed \cite{rfc8100}. Fifth, the burst scenarios are constructed and uncalibrated, and the real-harness measurement (Section~VI-L) used plain HTTP, non-streaming responses, an assumed model latency, synthetic background load, one congested router, and verification bypassed, in a single virtual machine. Finally, VATC is specified for edge and enterprise routers and not for core routers, and fairness among agentic flows was not evaluated.

\subsection{Future work}

Six next steps follow. (1) Search the \texttt{fq\_codel} parameter space, test per-flow fair queuing combined with a policed class (for example cake with DiffServ tins), and validate on ns-3 or a hardware testbed at production scale. (2) Measure real agent traffic to replace the synthetic and loopback assumptions. (3) Implement the gateway-to-router IPFIX channel, measure export-to-verdict latency on a hardware exporter, and evaluate the classifier on partial flows. (4) Repeat the delay measurement on physical networks with spoofers that adapt to the deployed verifier and a measured hop count, including loss and jitter. (5) Extend the AQI to multiple classes, design a DSCP-bleaching fallback, and replace the fixed cap with an adaptive or per-source policer. (6) Compare behavioral verification with signed or attested marking, and integrate attested agent identity \cite{aip} to address the compromised-agent threat.

\section{Conclusion}

This paper proposed the Verifiable Agentic Traffic Class VATC and evaluated it in two simulators and on a real Linux kernel. The results support a conditional conclusion. Where agent load was light or every sender was honest, giving agent traffic a class of its own in DiffServ did as well as VATC: steady-state latencies were identical, the honest kernel tests did not separate the two, and in two real agent harnesses separating bulk transfers from agent calls removed 96\% to 100\% of the queuing delay, whichever mechanism did it. In a simulated burst without a flood, a dedicated class cut mean agent queuing delay by 94\% to 98\% relative to DiffServ (the 98\% for VATC assumes an immediate verdict and is an upper bound) at a 12\% increase in human delay, so most of the gain came from isolating agent traffic.

The agent-specific parts addressed spoofed claims, and the rate limit bounded the harm of a spoofed flood to human traffic: human-surge latency was 25.4 ticks against 36.8 for the same verified design without the cap in the simulator at a flood four times the cap, and 0.059~ms against 0.65~ms under DiffServ in the kernel. Verification, with 89.2\% accuracy and a ROC-AUC of 0.954 on complete flows (15 seeds), leaked about 35\% of adversarial flows in the simulator. In the kernel, a live four-feature classifier demoted all 20 bulk-shaped spoofed sessions (median verdict 36~ms) and cut legitimate agent latency from 20.1 to 0.30~ms, below DiffServ's 0.62~ms and the tight \texttt{fq\_codel} setting's 3.80~ms, but it demoted none of six spoofers that copied the agent message pattern. Without verification, DiffServ and the tuned and tight \texttt{fq\_codel} settings gave legitimate agents lower latency than VATC under a flood, because the cap was sized for legitimate load and shared with the spoofed flows, but the tight \texttt{fq\_codel} setting raised human latency about 40-fold. The case for VATC therefore rests on the bound it places on human delay and, against spoofers whose behavior differs from an agent's, on verification.

VATC is worth considering on congested edge links where agent bursts coincide with human traffic, where senders may falsely claim agent status, and where a trusted gateway can supply the hop count; elsewhere a dedicated class is enough. The conclusions are limited by synthetic traffic, constructed bursts, sandbox-scale kernel tests, an assumed model latency, a generator-assigned hop count, and classifier scoring mostly on complete flows; under emulated wide-area delay, a classifier trained on local flows accepted every human and bulk flow at 80 and 150~ms until it was retrained. Next steps are measuring real agent traffic and repeating the kernel tests at production scale, building the gateway-to-router record channel and evaluating verification on partial flows and against adaptive spoofers under wide-area delay, and testing per-flow fair queuing combined with a policed class.

\section*{Data Availability}

The simulators, experiment scripts, the real-harness test package, and raw results are available in the artifact repository \cite{artifact}.

\section*{Acknowledgment}

The authors used Claude \cite{claude}, a large language model-based AI assistant from Anthropic, in preparing this manuscript and its revision, at a substantial level across the full text (Abstract through Conclusion) and across the paper's software: the primary and second simulators, the real-traffic measurement harness, the verification-classifier training code, and the real-kernel (Linux tc, HTB, Mininet) validation artifacts. The authors executed and verified these artifacts, including on external infrastructure outside the sandbox (Sections~VI-H to VI-J). The AI system also assisted in identifying and correcting methodological issues, including an error in the simulator's scheduler allocation, and in revising the manuscript in response to peer review. The authors reviewed and verified all claims, results, and conclusions and take full responsibility for them. No AI system is listed as an author, consistent with IEEE authorship policy.

\begin{IEEEbiography}[{\includegraphics[width=1in,height=1.25in,clip,keepaspectratio]{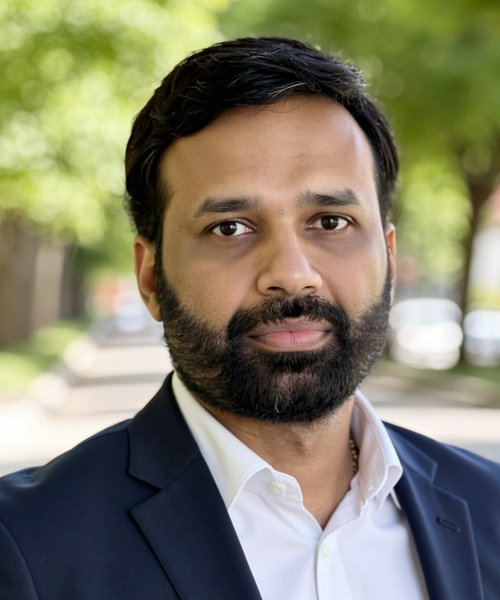}}]{Naveen Babu Narayanan Nair} is a Sr. Director of Software Development \& Engineering at Comcast Corporation, Philadelphia, PA, USA. His research interests include agentic network operations, LLM, AI, platform and tools, SaaS, automation frameworks, edge intelligence, and software-assisted optimization for Internet performance modeling. He received the Master of Computer Applications degree from Visvesvaraya Technological University. He is a Member of IEEE and ACM, and a mentor with Kerala Startup Mission. \newline Contact him at naveenbabunarayanannair@gmail.com.
\end{IEEEbiography}

\begin{IEEEbiography}[{\includegraphics[width=1in,height=1.25in,clip,keepaspectratio]{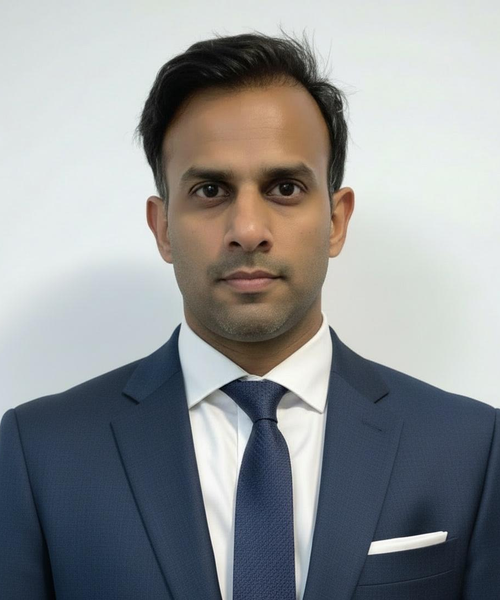}}]{Sajay Kurup} is a Sr. Manager of Software Development \& Engineering at Comcast Corporation, Philadelphia, PA, USA. His research interests include agentic network operations, edge intelligence, Internet performance modeling, and software-assisted optimization. He received the M.S. degree in Information Technology from Arizona State University. \newline Contact him at sajayjkurup@gmail.com. \EOD
\end{IEEEbiography}

\end{document}